\documentclass{article}

\PassOptionsToPackage{numbers, compress}{natbib}
\usepackage[preprint]{neurips_2026}

\usepackage[utf8]{inputenc} 
\usepackage[T1]{fontenc}    
\usepackage{hyperref}       
\usepackage{url}            
\usepackage{booktabs}       
\usepackage{amsfonts}       
\usepackage{nicefrac}       
\usepackage{microtype}      
\usepackage{xcolor}         
\usepackage{amsmath}
\usepackage{booktabs}
\usepackage[table]{xcolor}
\usepackage{graphicx}
\usepackage{booktabs}
\usepackage{multirow}
\usepackage{makecell}
\usepackage{subcaption}
\usepackage{wrapfig}

\title{Retrieve-then-Steer: Online Success Memory for Test-Time Adaptation of Generative VLAs}

\author{
Jianchao Zhao\textsuperscript{1,2},
Huoren Yang\textsuperscript{1,2},
Yusong Hu\textsuperscript{2},
Yuyang Gao\textsuperscript{2},
Qiguan Ou\textsuperscript{2},\\
Cong Wan\textsuperscript{1},
SongLin Dong\textsuperscript{3},
Zhiheng Ma\textsuperscript{3},
Yihong Gong\textsuperscript{1,3},
\\[0.8em]
\textsuperscript{1}College of Artificial Intelligence, Xi'an Jiaotong University\\
\textsuperscript{2}One Robotics\\
\textsuperscript{3}Shenzhen University of Advanced Technology\\
}

\begin{document}

\maketitle

\begin{abstract}
Vision-Language-Action (VLA) models show strong potential for general-purpose robotic manipulation, yet their closed-loop reliability often degrades under local deployment conditions. Existing evaluations typically treat test episodes as independent zero-shot trials. However, real robots often operate repeatedly in the same or slowly changing environments, where successful executions provide environment-verified evidence of reliable behavior patterns. We study this persistent-deployment setting, asking whether a partially competent frozen VLA can improve its reliability by reusing its successful test-time experience. We propose an online success-memory guided test-time adaptation framework for generative VLAs. During deployment, the robot stores progress-calibrated successful observation-action segments in a long-term memory. At inference, it retrieves state-relevant action chunks, filters inconsistent candidates via trajectory-level consistency, and aggregates them into an elite action prior. To incorporate this prior into action generation, we introduce confidence-adaptive prior guidance, which injects the elite prior into an intermediate state of the flow-matching action sampler and adjusts the guidance strength based on retrieval confidence. This design allows the frozen VLA to exploit environment-specific successful experience while preserving observation-conditioned generative refinement. This retrieve-then-steer mechanism enables lightweight, non-parametric test-time adaptation without requiring parameter updates. Simulation and real-world experiments show improved task success and closed-loop stability, especially in long-horizon and multi-stage tasks.

\end{abstract}

\section{Introduction}


Vision-Language-Action (VLA) models~\citep{rt-1,rt-2,openvla,pi_0}, particularly generative VLAs~\citep{diffusion,pi05,cogact} that incorporate diffusion or flow-matching mechanisms, show immense potential for general-purpose robotic manipulation by generating expressive and temporally coherent action chunks. However, a significant disconnect exists between current evaluation paradigms and real-world deployment. Most benchmarks treat testing as independent zero-shot trials, overlooking that real robots typically perform repetitive tasks in static or slowly changing environments. In such settings, physical layouts, camera viewpoints, calibration errors, and task patterns exhibit strong inter-episode correlation. Consequently, deployment should not be viewed as a series of isolated test episodes, but rather as a persistent online process operating under correlated local conditions.


Embracing this "persistent online" perspective is vital. While many existing VLAs possess strong foundational capabilities, their closed-loop execution often remains unstable during real-world deployment. Although a robot might occasionally complete a task, it is highly prone to failure in nearly identical states due to perception noise, viewpoint shifts, or accumulated errors~\cite{Libero-plus,romer2025failure}. This fragility underscores the value of successful experiences. A successful grasp or placement, for instance, implicitly captures the visual geometry, actuation biases, and execution timing specific to that environment. Consequently, these trajectories should not be treated as isolated samples discarded after evaluation, but rather as environment-verified evidence that dictates reliable behavior patterns under the current physical and visual settings. This motivates our central question: \textit{can a frozen base VLA improve its reliability by reusing its own successful test-time interactions?}


\begin{wrapfigure}{r}{0.6\linewidth}
    \vspace{-0.8em}
    \centering
    \includegraphics[width=\linewidth]{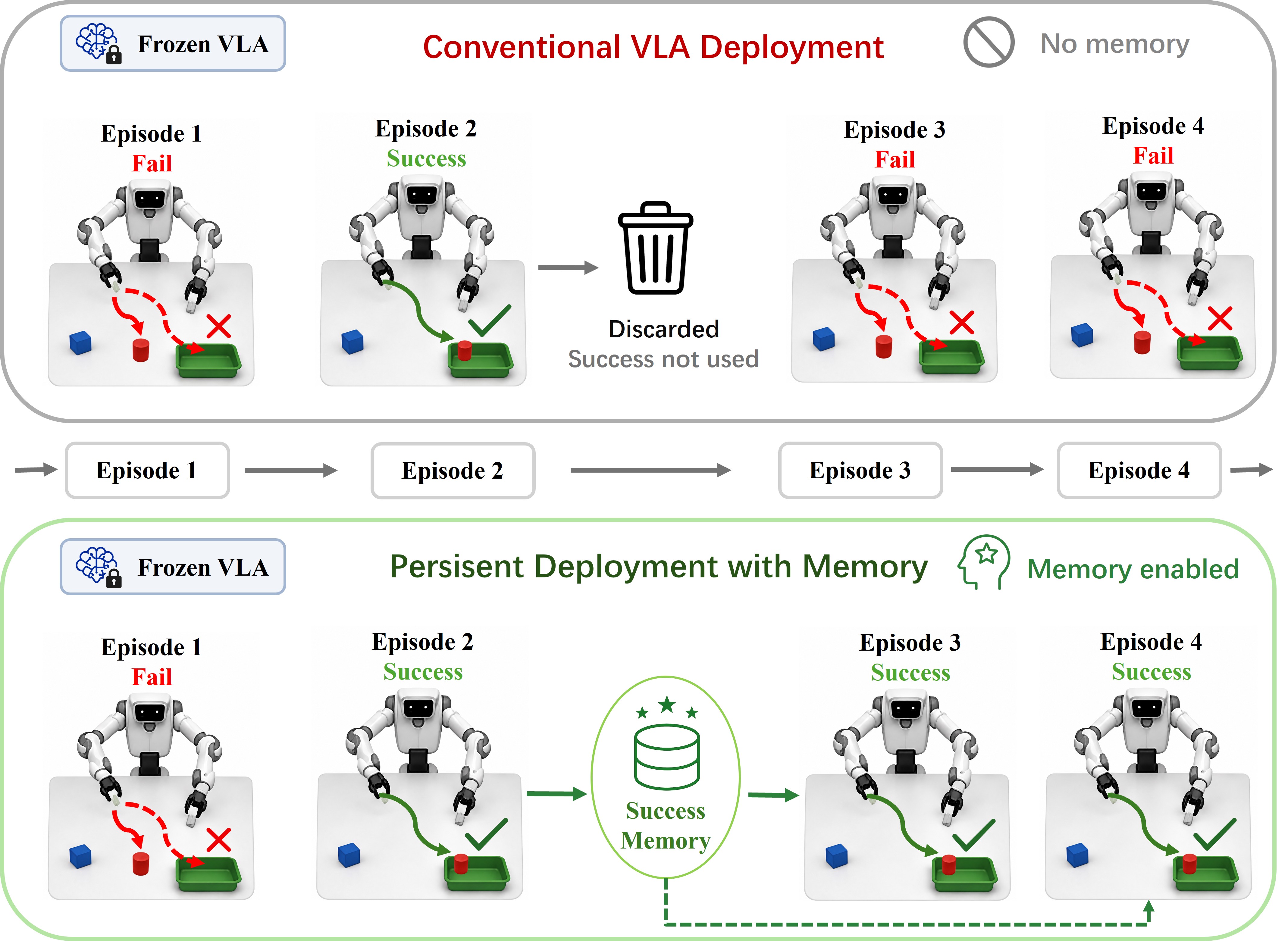}
    \caption{
    Motivation for persistent VLA deployment. 
    Our method stores successful trials in online memory and reuses them to stabilize later executions without updating the policy.
    }
    \label{fig:motivation}
    \vspace{-1.0em}
\end{wrapfigure}

A review of existing research, however, reveals that current paradigms have yet to provide a satisfactory answer to this question. First, pre-training and downstream fine-tuning~\citep{Open-x-embodiment,openvla,pi05} enhance policies before deployment but fail to enable continuous learning from test-time successes. To this end, reinforcement learning and human-in-the-loop methods~\citep{Hg-dagger,reduction_rl} utilize deployment experience but typically require extra feedback, safe exploration, and heavy parameter updates. In contrast, recent test-time steering~\citep{taco,robomonkey,Verifier-free,rover} focuses on inference by sampling and filtering action candidates. However, these training-free approaches generally follow a myopic "generate-then-select" paradigm, discarding candidates after local evaluation. Consequently, they struggle to exploit successful experiences accumulated across repeated episodes.



Therefore, we propose directly formulating successful test-time experience as a soft behavioral prior for generative VLAs. Rather than assuming independent test episodes, we organize successful executions as reusable local evidence for future action generation. This yields a novel "retrieve-then-steer" paradigm: an online success memory compels the frozen VLA toward behavior patterns proven effective in the target environment, while preserving its ability to condition on real-time observations.


To operationalize this paradigm, we introduce an online success-memory guided test-time adaptation framework for generative VLAs. 
Specifically, during continuous deployment, the robot stores progress-calibrated successful observation-action prefixes while excluding failed or redundant motions. At inference, it retrieves observation-relevant action chunks, eliminates conflicting trajectories via consistency filtering, and aggregates high-quality candidates into an elite action prior. To integrate this prior into action generation, we propose confidence-adaptive prior guidance. By injecting the elite prior directly into the intermediate state of the generative sampler, the system dynamically adjusts guidance strength based on retrieval confidence. This ensures high-confidence retrievals dictate successful behavior patterns, while uncertain retrievals safely revert to the original VLA sampler.


We evaluate the proposed framework in both \textbf{simulation and real-world} robotic manipulation. Our method improves generative VLA policies on long-horizon language-conditioned manipulation benchmarks, including LIBERO-10 and SimplerEnv, and further shows consistent gains on real-world bimanual manipulation tasks. Across these settings, online success-memory guidance improves task success and closed-loop stability, especially on long-horizon and multi-stage tasks. Our contributions are threefold:




\begin{itemize}
    \item 
    We redefine VLA deployment as a persistent online adaptation process rather than isolated trials, highlighting successful test-time interactions as a critical source of environment-specific evidence for enhancing policy reliability.
    \item We propose a non-parametric "retrieve-then-steer" mechanism that enables lightweight TTA for frozen VLAs. This mechanism utilizes a progress-calibrated success memory to extract reusable segments and injects consistency-filtered elite priors into the generative sampling process, guiding the model without requiring parameter updates.
    \item We systematically validate the framework across long-horizon benchmarks and real-world bimanual manipulation. Results demonstrate our method significantly improves success rates and strengthens closed-loop stability in complex, multi-stage tasks.
\end{itemize}
\section{Related Works}
\paragraph{Vision-Language-Action Models.}
Vision-Language-Action models (VLAs)~\citep{rt-1,rt-2,openvla,Open-x-embodiment,dita} have become a promising paradigm for general-purpose robotic policies by unifying visual perception, language understanding, and action generation. Early systems such as RT-1~\citep{rt-1} and OpenVLA~\citep{openvla} learn end-to-end policies from large-scale robotic data, while recent generative policies, including Diffusion Policy~\citep{diffusion} and $\pi_0/\pi_{0.5}$~\citep{pi_0,pi05}, model continuous action chunks with diffusion or flow-matching heads. Despite these advances, VLAs still suffer from sampling noise, distribution shifts, and accumulated closed-loop errors during deployment, limiting their stability and local adaptability.
\paragraph{Test-Time Policy Steering.}
Recent work has explored test-time policy steering or scaling to improve VLA deployment stability~\citep{Verifier-free,rover,robomonkey,taco}. These methods enhance current action decisions through additional sampling, external evaluators, or internal confidence signals. For example, RoboMonkey~\citep{robomonkey} selects among perturbed action candidates with a VLM-based verifier, MG-Select~\citep{Verifier-free} uses condition-masking confidence for verifier-free selection, and TACO~\citep{taco} constrains generation toward stable successful modes via pseudo-count estimation. While effective, these methods follow a generate-then-select paradigm, which incurs extra inference overhead and discards reusable cross-episode experience. By contrast, our method performs prior-guided generation, retrieving successful action segments to steer the generative sampler before actions are produced.
\paragraph{Retrieval-Augmented and Memory-Based Robot Learning.}
Retrieval-augmented and memory-based mechanisms~\citep{memoryvla,strap,retrieval-skill-base} have long been used in robot learning to improve the utilization of historical experience, demonstrations, and task context. Existing methods typically retrieve relevant trajectories from offline demonstration datasets for few-shot imitation, skill retrieval, or local policy adaptation~\citep{strap,retrieval,expres-vla}, while others treat memory as a replay buffer for continual learning or reinforcement learning~\citep{continually,conrft}. Although effective, these methods often rely on offline data, use retrieval mainly as context, or require policy updates, limiting their suitability for lightweight test-time adaptation of frozen VLAs. Unlike these approaches, our method constructs memory directly during deployment from verified successful executions and uses it as a lightweight non-parametric prior for frozen VLAs, without offline demonstration banks or parameter updates.
\section{Preliminaries} 
\subsection{Problem Formulation}
We consider language-conditioned robotic manipulation in downstream deployment. Let $\pi_{\mathrm{vla}}$ be a generative Vision-Language-Action policy fine-tuned on downstream demonstrations. At decision step $t$, given observation $o_t=(I_t^{1:N_c},q_t)$ and instruction $l$, the policy samples an action chunk $a_t \sim \pi_{\mathrm{vla}}(\cdot \mid o_t,l)$ with horizon $H$, where $I_t^{1:N_c}$ denotes multi-view RGB images and $q_t$ denotes the proprioceptive state.
During each test episode, the robot executes action chunks in closed loop, producing a trajectory $\tau^{(i)}=\{(o_t^{(i)},a_t^{(i)})\}_{t=0}^{T_i-1}$. Unlike standard zero-shot evaluation that treats episodes independently, we study continuous deployment, where successful cross-episode experience can be accumulated and reused for test-time adaptation.

\begin{figure*}[t]
    \centering
    \includegraphics[width=1.0\textwidth]{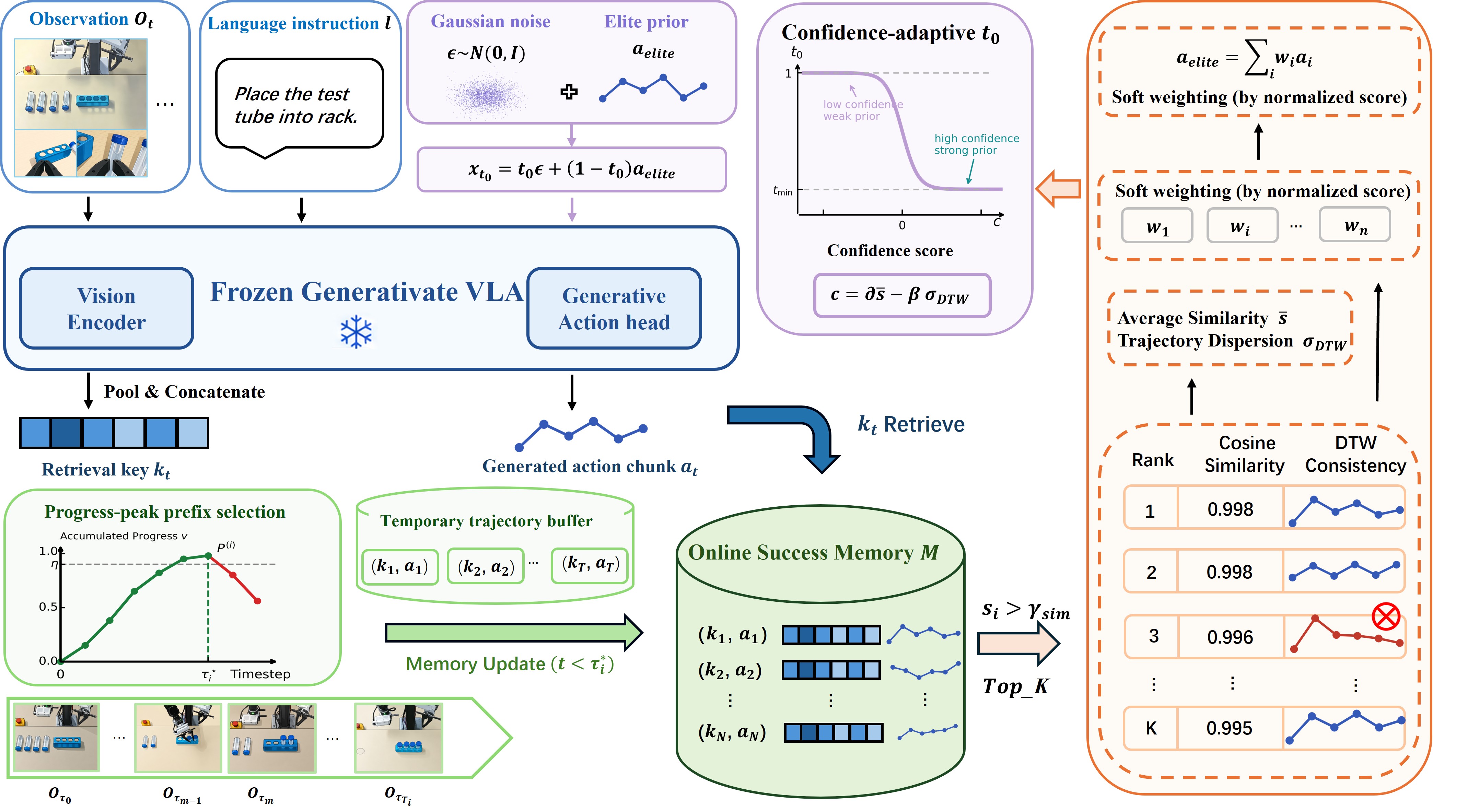}
    \caption{Overview of our retrieve-then-steer test-time adaptation framework. The frozen VLA accumulates progress-verified successful observation--action segments in an online success memory. For each new observation, relevant action chunks are retrieved, filtered, and aggregated into an elite prior, which initializes the flow-matching sampler with confidence-adaptive guidance.
    }
    \label{fig:main}
\end{figure*}

\section{Methodology}
\subsection{Online Progress-Calibrated  Memory}
Since the frozen VLA cannot update its parameters during deployment, we construct an Online Progress-Calibrated Memory \(\mathcal{M}\) to store successful observation--action segments.
\paragraph{Trajectory buffering and memory representation.}
For the $i$-th test episode, we maintain a temporary buffer $\mathcal{B}^{(i)}=\{(k_t^{(i)},a_t^{(i)})\}_{t=0}^{T_i-1}$  to record candidate memory entries generated during execution. Here, $k_t^{(i)}$ denotes the retrieval key at time step $t$. Instead of introducing a separate retrieval encoder, we reuse the VLA visual encoder to extract image features. For each view, spatial patch tokens are reshaped into a feature grid, downsampled by $2\times2$ average pooling, and flattened. For multi-view observations, features from all views are concatenated and normalized to form $k_t^{(i)}$.
\paragraph{Interval-based progress calibration.}
To identify reusable successful experience, we instantiate the progress estimator $\Phi_\psi$ with a pretrained VLAC critic~\citep{vlac}, which automatically evaluates trajectories for memory construction without human success labels. For each task, one successful demonstration video from the training set is used as the reference process $R$, serving as an in-context example of the task execution procedure. Details are provided in Appendix~\ref{app:progress_estimator}.

Let $\Delta$ denote the evaluation interval. For trajectory $i$, we evaluate progress at timesteps 
$\mathcal{T}_i^\Delta=\{0,\Delta,2\Delta,\ldots,\lfloor T_i/\Delta \rfloor\Delta\}\cup\{T_i\}$.
For adjacent timesteps $\tau_{m-1},\tau_m\in\mathcal{T}_i^\Delta$, the pretrained critic predicts the signed progress change conditioned on the instruction and reference process:
\begin{equation}
c_{\tau_m}^{(i)}
=
\Phi_\psi
\left(
o_{\tau_{m-1}}^{(i)},
o_{\tau_m}^{(i)},
l^{(i)}; R
\right),
\end{equation}
the value of $c_{\tau_m}^{(i)}$ indicates whether the task progresses or regresses within the interval. We then accumulate interval-level progress into a trajectory-level progress score $v_{\tau_m}^{(i)}$, initialized as $v_{\tau_0}^{(i)}=0$:
\begin{equation}
v_{\tau_m}^{(i)}
=
v_{\tau_{m-1}}^{(i)}
+
\left(100-v_{\tau_{m-1}}^{(i)}\right)
\frac{c_{\tau_m}^{(i)}}{100}.
\end{equation}

\paragraph{Progress-peak prefix selection and memory update.}
After the episode finishes, we define the completion score as the maximum accumulated progress,
$P^{(i)}=\max_{\tau_m\in\mathcal{T}_i^\Delta} v_{\tau_m}^{(i)}$,
and denote the progress-peak timestep as
$\tau_i^\star=\arg\max_{\tau_m\in\mathcal{T}_i^\Delta} v_{\tau_m}^{(i)}$. We use the maximum accumulated progress instead of terminal progress to handle possible regressions after near-success, such as overshooting, collisions, or unnecessary motions. This allows the progress peak to preserve the best achieved task state and retain reusable successful experience. Given a success threshold $\eta$, the episode-level success indicator is defined as
$y^{(i)}=\mathbb{I}[P^{(i)}\ge\eta]$.
If $y^{(i)}=0$, the temporary buffer is discarded; otherwise, we retain only candidate entries before the progress peak:
$\mathcal{B}_{+}^{(i)}=\{(k_t^{(i)},a_t^{(i)})\in\mathcal{B}^{(i)}\mid t\le\tau_i^\star\}$. Then the online success memory is updated as $\mathcal{M}\leftarrow \mathcal{M}\cup \mathcal{B}_{+}^{(i)}$.

\subsection{Retrieval-based Action Prior}
When the online success memory $\mathcal{M}$ is non-empty, we retrieve historical successful actions that are similar to the current state during test time, which are used to assist subsequent action generation.
\paragraph{Successful action retrieval with similarity gating.}
Given the current retrieval key $k_t$, for each memory entry $(k_i,a_i)\in\mathcal{M}$, we compute its relevance to the current state using cosine similarity, $s_i=\frac{\langle k_t,k_i\rangle}{\|k_t\|_2\|k_i\|_2}$. We then select the top-$K$ candidates with the highest similarity scores and remove weakly related results using a threshold $\gamma_{\mathrm{sim}}$, yielding the initial candidate set $\mathcal{I}_{\mathrm{sim}}$.

\paragraph{DTW-based trajectory consistency filtering.}
State-level similarity alone may still introduce action-level mismatches, where
retrieved states are close to the current state but their action chunks follow
inconsistent trajectory patterns. To remove such outliers, we compute pairwise
multivariate Dynamic Time Warping (DTW) distances among the candidates in
\(\mathcal{I}_{\mathrm{sim}}\):
\begin{equation}
d_{ij}
=
\mathrm{DTW}(a_i, a_j),
\quad
i,j\in\mathcal{I}_{\mathrm{sim}}.
\end{equation}
For each candidate action chunk $a_i$, we define its trajectory inconsistency score as the median distance to the remaining candidates, 
$r_i=\mathrm{median}_{j\in\mathcal{I}_{\mathrm{sim}},\, j\neq i} d_{ij}$.
A larger \(r_i\) indicates that the candidate deviates from the dominant
successful trajectory pattern. We therefore remove candidates with excessively
large inconsistency scores and obtain the final candidate set \(\mathcal{I}\).

\paragraph{Elite action prior aggregation.}
Given the filtered candidate set $\mathcal{I}$, we aggregate multiple
successful action chunks with similarity-based soft weights, rather than
directly selecting a single nearest-neighbor action. The weight of each
candidate is defined as
\begin{equation}
w_i =
\frac{
\exp\left((s_i-\max_{j\in\mathcal{I}}s_j)/\tau\right)
}{
\sum_{j\in\mathcal{I}}
\exp\left((s_j-\max_{m\in\mathcal{I}}s_m)/\tau\right)
},
\quad i\in\mathcal{I},
\end{equation}
where $\tau>0$ is a temperature parameter that controls the sharpness of the
weight distribution. The resulting elite action prior is given by
\begin{equation}
a_{\mathrm{elite}}
=
\sum_{i\in\mathcal{I}} w_i a_i .
\end{equation}
This formulation provides a unified representation for action-prior aggregation. For action components in Euclidean spaces, such as positions, joint angles, we adopt linear weighted aggregation; for orientations, we compute the geodesic mean on $SO(3)$. Further details are provided in Appendix~\ref{app:component_aggregation}.


\subsection{Confidence-Adaptive Prior Guidance}
The retrieved elite action prior $a_{\mathrm{elite}}$ provides a local successful behavior reference from the online success memory. However, unreliable retrievals caused by representation bias, nearest-neighbor mismatch, or trajectory inconsistency may introduce incorrect constraints. We therefore propose \emph{confidence-adaptive prior guidance}, which injects the retrieved prior into the flow-matching sampler as a soft generative constraint and adapts its strength according to retrieval confidence.

For a VLA with a flow-matching action head, the original sampler starts from Gaussian noise $x_1=\epsilon,\epsilon\sim\mathcal{N}(0,I)$ and integrates the conditional velocity field $v_\theta(x_t,t,z_t)$ from $t=1$ to $t=0$, where $z_t$ is the conditioning feature from the current observation and instruction. Instead of modifying the model or velocity field, we initialize the sampling process from an intermediate state:
\begin{equation}
x_{t_0}
=
(1-t_0)a_{\mathrm{elite}}
+
t_0\epsilon,
\qquad
\epsilon\sim\mathcal{N}(0,I).
\end{equation}
Here, $t_0\in[0,1]$ controls the guidance strength. A smaller $t_0$ places the initial state closer to $a_{\mathrm{elite}}$, yielding stronger prior guidance, while a larger $t_0$ preserves more randomness and recovers the original sampler.

To adapt the guidance strength to retrieval reliability, we estimate a confidence score from both state-level similarity and action-level consistency. 
Given the filtered candidate set $\mathcal{I}$, we compute the average retrieval similarity 
$\bar{s}_{\mathrm{top}\text{-}K}=\frac{1}{|\mathcal{I}|}\sum_{i\in\mathcal{I}}s_i$. 
Since deployment similarities often lie in a narrow high-score range, we normalize it as
\begin{equation}
\tilde{s}
=
\mathrm{clip}
\left(
\frac{\bar{s}_{\mathrm{top}\text{-}K}-s_{\mathrm{ref}}}{s_{\mathrm{scale}}},
-c_{\max},
c_{\max}
\right).
\end{equation}
We further measure action-level dispersion using the DTW inconsistency scores,
$\sigma_{\mathrm{DTW}}=\mathrm{Std}(\{r_i\}_{i\in\mathcal{I}})$, and define the retrieval confidence as
\begin{equation}
c
=
\alpha \tilde{s}
-
\beta \sigma_{\mathrm{DTW}},
\end{equation}
where larger $c$ indicates a more reliable prior. Finally, we map the retrieval confidence to the sampling starting time:
\begin{equation}
t_0
=
t_{\min}
+
(1-t_{\min})
\sigma(-\gamma c),
\end{equation}
where $t_{\min}$ is the strongest-guidance starting time and $\gamma$ controls the mapping sharpness. 
Higher confidence moves $t_0$ toward $t_{\min}$, while lower confidence moves it toward $1$, recovering the original sampler.

After determining $t_0$, the sampler starts from $x_{t_0}$ and integrates the same conditional velocity field to $t=0$. With Euler discretization, the update is
\begin{equation}
x_{t-\Delta t}
=
x_t
-
\Delta t \cdot v_\theta(x_t,t,z_t),
\qquad
t:t_0\rightarrow 0 .
\end{equation}
The final state $x_0$ serves as the generated action chunk $\hat{a}_{t:t+H-1}$. If the memory is empty or retrieval fails the similarity and trajectory-consistency filters, no prior is injected and and the method falls back to the original sampling process. We also provide the diffusion counterpart in Appendix~\ref{app:diffusion_prior_guidance}.

\section{Experiments}
\subsection{Simulation Experiments}
\subsubsection{Setup and Baselines}
\paragraph{Benchmarks.}
We evaluate our method on two simulation benchmarks, LIBERO\citep{libero} and SimplerEnv\citep{simpler}. LIBERO is a benchmark for lifelong learning in decision making, consisting of multiple task suites. As easier suites are near-saturated, we focus on the more challenging LIBERO-10 suite to examine whether our method can mitigate state drift, accumulated action errors, and unstable closed-loop execution in long-horizon, multi-stage manipulation tasks. SimplerEnv is a real-to-sim manipulation benchmark built upon the SAPIEN simulator and the ManiSkill2 benchmark, providing simulated task environments for both the WidowX and Google Robot platforms. In this work, we primarily use the tasks designed for the Google Robot platform to evaluate robustness under realistic deployment conditions, including object layout variations, visual perturbations, and fine-grained manipulation. 

\paragraph{Baselines.}

\begin{table}[t]
\centering
\caption{Success rates (\%) on LIBERO-10. Each task is evaluated over 50 trials. \textsuperscript{*} denotes reproduced results. For reproduced results, the average row reports mean $\pm$ std over three random seeds. }
\label{tab:libero10_results}
\small
\setlength{\tabcolsep}{3.5pt}
\renewcommand{\arraystretch}{1.08}
\resizebox{\textwidth}{!}{
\begin{tabular}{lccccccc}
\toprule
\textbf{Task}

& OpenVLA~\citep{openvla}
& $\pi_0$-FAST~\citep{pi0fast}
& $\pi_0$\textsuperscript{*}~\citep{pi_0}
& $\pi_0$ + TACO\textsuperscript{*}~\citep{taco}
& \cellcolor{gray!10}$\pi_0$ + Ours
& $\pi_{0.5}$\textsuperscript{*}~\citep{pi05}
& \cellcolor{gray!10}$\pi_{0.5}$ + Ours \\
\midrule

Soup and Sauce in Basket
& 60.0 & 74.0 & 78.0 & 82.0 & \cellcolor{gray!10}84.0 & 90.0 & \cellcolor{gray!10}\textbf{100.0} \\

Cheese and Butter in Basket
& 76.0 & 72.0 & 98.0 & 94.0 & \cellcolor{gray!10}92.0 & \textbf{100.0} & \cellcolor{gray!10}\textbf{100.0} \\

Turn on Stove and Place Moka
& 58.0 & 62.0 & 84.0 & 92.0 & \cellcolor{gray!10}96.0 & 96.0 & \cellcolor{gray!10}\textbf{98.0} \\

Black Bowl in Drawer
& 36.0 & 52.0 & 90.0 & 92.0 & \cellcolor{gray!10}96.0 & 94.0 & \cellcolor{gray!10}\textbf{100.0} \\

Mugs on Plates
& 32.0 & 54.0 & 84.0 & 82.0 & \cellcolor{gray!10}82.0 & \textbf{96.0} & \cellcolor{gray!10}\textbf{96.0} \\

Book in Caddy
& 82.0 & 82.0 & 96.0 & 94.0 & \cellcolor{gray!10}96.0 & \textbf{100.0} & \cellcolor{gray!10}92.0 \\

Mug and Pudding on Plate
& 60.0 & 58.0 & 82.0 & 82.0 & \cellcolor{gray!10}80.0 & \textbf{94.0} & \cellcolor{gray!10}\textbf{94.0} \\

Soup and Cheese in Basket
& 70.0 & 72.0 & 98.0 & 96.0 & \cellcolor{gray!10}94.0 & 96.0 & \cellcolor{gray!10}\textbf{100.0} \\

Moka Pots on Stove
& 20.0 & 26.0 & 30.0 & 36.0 & \cellcolor{gray!10}38.0 & 64.0 & \cellcolor{gray!10}\textbf{70.0} \\

Mug in Microwave
& 46.0 & 50.0 & 76.0 & 88.0 & \cellcolor{gray!10}86.0 & \textbf{94.0} & \cellcolor{gray!10}\textbf{94.0} \\


\midrule
Average
& \makecell{54.0 \\ {\scriptsize --}}
& \makecell{60.2 \\ {\scriptsize --}}
& \makecell{81.6 \\ {\scriptsize $\pm$ 0.8}}
& \makecell{83.8 {\scriptsize $(\uparrow 2.2)$} \\ {\scriptsize $\pm$ 0.2}}
& \cellcolor{gray!10}\makecell{84.4 {\scriptsize $(\uparrow 2.8)$} \\ {\scriptsize $\pm$ 0.4}}
& \makecell{92.4 \\ {\scriptsize $\pm$ 0.2}}
& \cellcolor{gray!10}\makecell{\textbf{94.4} {\scriptsize $(\uparrow 2.0)$} \\ {\scriptsize $\pm$ 0.3}} \\
\bottomrule

\end{tabular}
}
\end{table}

We mainly evaluate our framework on VLA policies with flow-matching or diffusion-based action heads. 
Specifically, we select $\pi_0$\citep{pi_0} and $\pi_{0.5}$\citep{pi05} as the primary baseline models for LIBERO, and adopt CogACT\citep{cogact} for experiments on SimplerEnv. We also compare with TACO~\citep{taco}, a test-time scaling method, to evaluate our method against existing test-time steering approaches. For a more comprehensive comparison, we further report the success rates of representative VLA policies on selected benchmarks, including OpenVLA\citep{openvla}, $\pi_0$-FAST~\citep{pi0fast}, RT-1\citep{rt-1}, RT-1-X\citep{Open-x-embodiment}, RT-2-X\citep{Open-x-embodiment}, and Octo\citep{octo}.



\subsubsection{Results}
The simulation results are reported in Tables~\ref{tab:libero10_results} and~\ref{tab:simpler_results}. On LIBERO-10, our method improves both base policies. 
For $\pi_0$, the average success rate increases from 81.6\% to 84.4\%, outperforming the test-time scaling method TACO, which achieves 83.8\%. 
For the stronger $\pi_{0.5}$ policy, our method further improves the average success rate from 92.4\% to 94.4\%. Task-level gains are especially clear on long-horizon and multi-stage tasks such as Turn on Stove and Place Moka, Black Bowl in Drawer, Moka Pots on Stove, and Soup and Sauce in Basket. These results show that online success memory provides reusable environment-specific action priors that help stabilize closed-loop execution.

\begin{table}[t]
\centering
\caption{Success rates (\%) on the SIMPLER benchmark. We compare our method on top of CogACT with prior VLA policies. For CogACT and CogACT + Ours, we report mean $\pm$ std over three random seeds.}
\label{tab:simpler_results}
\setlength{\tabcolsep}{4.5pt}
\footnotesize
\renewcommand{\arraystretch}{1.05}
\resizebox{0.95\textwidth}{!}{
\begin{tabular}{lccccc}
\toprule
\textbf{Method}
& \makecell{\textbf{Pick}\\\textbf{Coke Can}}
& \makecell{\textbf{Move}\\\textbf{Near}}
& \makecell{\textbf{Open/Close}\\\textbf{Drawer}}
& \makecell{\textbf{Open Top Drawer}\\\textbf{and Place Apple}}
& \textbf{\textit{Average}} \\
\midrule
RT-1~\citep{rt-1}
& 85.7 & 44.2 & 73.0 & 6.5 & 52.4 \\

RT-1-X~\citep{Open-x-embodiment}
& 56.7 & 31.7 & 59.7 & 21.3 & 42.4 \\

RT-2-X~\citep{Open-x-embodiment}
& 78.7 & 77.9 & 25.0 & 3.7 & 46.3 \\

Octo-Base~\citep{octo}
& 17.0 & 4.2 & 22.7 & 0.0 & 11.0 \\

OpenVLA~\citep{openvla}
& 18.0 & 56.3 & 63.0 & 0.0 & 34.3 \\

CogACT~\citep{cogact}
& 91.3 {\scriptsize $\pm$ 0.3}
& 83.3 {\scriptsize $\pm$ 0.6}
& 71.8 {\scriptsize $\pm$ 0.2}
& 56.8 {\scriptsize $\pm$ 0.1}
& 75.8 {\scriptsize $\pm$ 0.3} \\

\rowcolor{gray!10}
\textbf{CogACT + Ours}
& \textbf{94.6} {\scriptsize $\pm$ 0.2} {\scriptsize $(\uparrow 3.3)$}
& \textbf{85.8} {\scriptsize $\pm$ 0.2} {\scriptsize $(\uparrow 2.5)$}
& \textbf{75.4} {\scriptsize $\pm$ 0.3} {\scriptsize $(\uparrow 3.6)$}
& \textbf{62.3} {\scriptsize $\pm$ 0.2} {\scriptsize $(\uparrow 5.5)$}
& \textbf{79.5} {\scriptsize $\pm$ 0.2} {\scriptsize $(\uparrow 3.7)$} \\

\bottomrule
\end{tabular}
}
\end{table}
On SimplerEnv, our method also improves CogACT from 75.8\% to 79.5\% on average, with consistent gains across all four tasks. The improvements are 3.3, 2.5, 3.6, and 5.5 points on Pick Coke Can, Move Near, Open/Close Drawer, and Open Top Drawer and Place Apple, respectively. The largest gain appears on the most challenging long-horizon task, suggesting that our retrieval-guided prior is effective under layout variations and visual perturbations.
~\begin{table}[t]
\centering
\caption{Performance comparison on the real-robot test tube placement task. The task is to pick up the test tubes one by one from left to right and place them onto the test tube rack.}
\label{tab:real_robot_test_tube}
\small
\begin{tabular*}{0.9\textwidth}{@{\extracolsep{\fill}}lccccc}
\toprule
\textbf{Method}
& \multicolumn{4}{c}{\textbf{Success Rate (\%)}}
& \textbf{Avg. Len} \\
\cmidrule(lr){2-5}
& \textbf{1/4}
& \textbf{2/4}
& \textbf{3/4}
& \textbf{4/4}
&  \\
\midrule
$\pi_0$~\citep{pi_0}
& 64.0 & 26.0 & 14.0 & 8.0 & 1.12 \\
$\pi_{0.5}$~\citep{pi05}
& 80.0 & 32.0 & 24.0 & 18.0 & 1.54 \\
$\pi_{0.5}$ + Ours
& 90.0 & 48.0 & 32.0 & 24.0 & 1.94 \\
\bottomrule
\end{tabular*}
\end{table}
~
\begin{table}[t]
\centering
\caption{
Component ablation on LIBERO-10. We report the average success rate (\%) over all tasks.
The ablation separately studies how the retrieved prior is constructed and how it is used for action generation.
All intermediate-initialization variants use dynamic $t_0$.
}
\scriptsize
\label{tab:component_ablation}
\setlength{\tabcolsep}{4.5pt}
\renewcommand{\arraystretch}{1.05}
\resizebox{0.8\textwidth}{!}{
\begin{tabular}{llcc}
\toprule
\textbf{Variant}
& \textbf{Retrieved Prior}
& \textbf{Prior Usage}
& \textbf{Avg. Success} \\
\midrule
Base $\pi_{0.5}$
& --
& Original sampler
& 92.4 \\

\midrule
\multicolumn{4}{l}{\textit{Prior-construction ablation}} \\

Top-1 Retrieval
& Nearest success chunk
& Intermediate init.
& 93.6 \\

Top-$K$ Soft Aggregation
& Top-$K$ weighted prior
& Intermediate init.
& 94.0 \\

\midrule
\multicolumn{4}{l}{\textit{Prior-usage ablation}} \\

Direct Replay
& Top-$K$ + DTW prior
& Direct execution
& 87.8 \\

Output Interpolation
& Top-$K$ + DTW prior
& Post-hoc interpolation
& 93.0 \\

\midrule
Full Ours
& Top-$K$ + DTW prior
& Intermediate init.
& 94.4 \\

\bottomrule
\end{tabular}
}
\end{table}

\begin{figure}[t]
    \centering
    \begin{subfigure}[t]{0.45\textwidth}
        \centering
        \includegraphics[width=\linewidth]{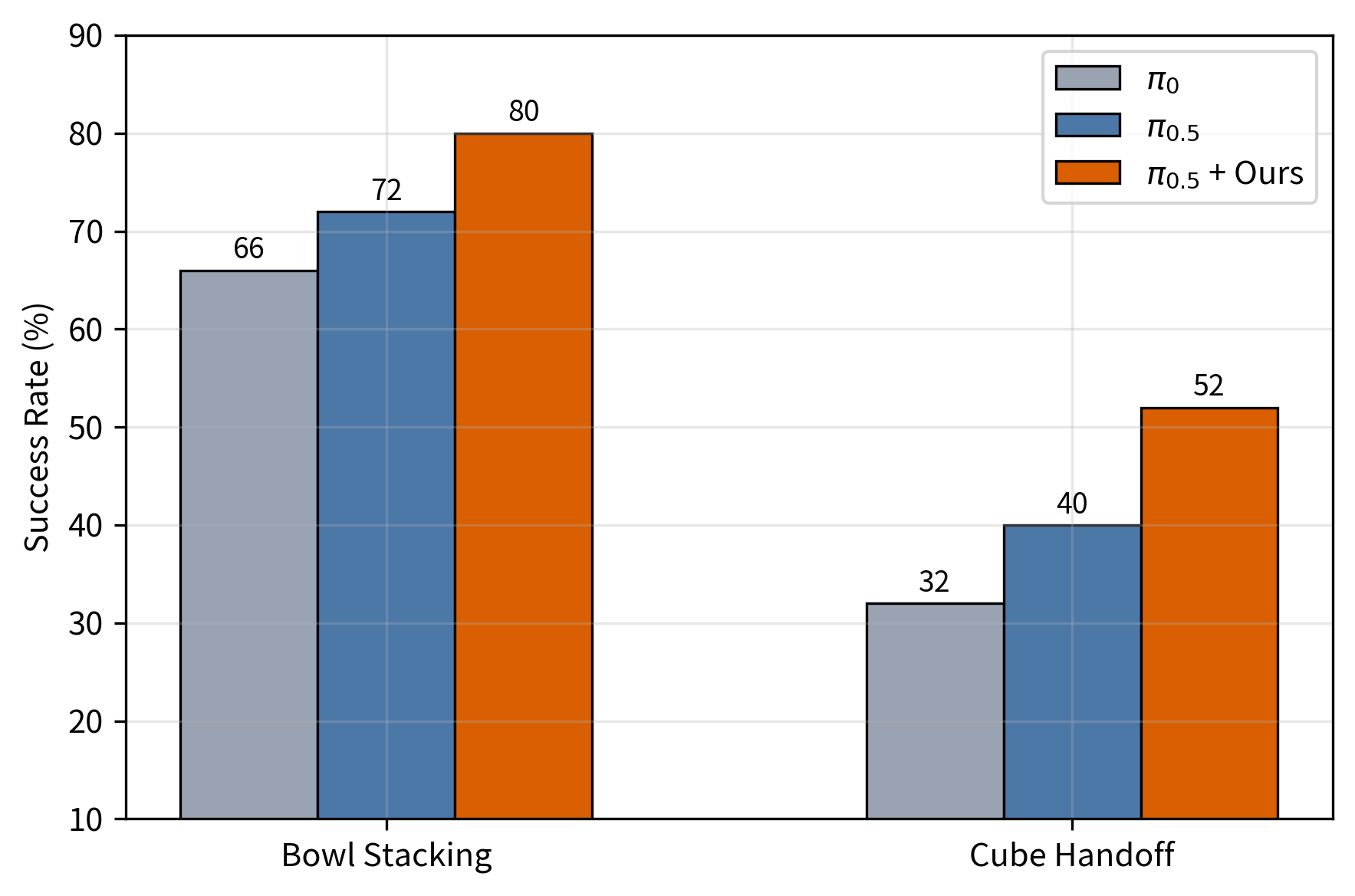}
        \caption{OpenArm tasks.}
        \label{fig:real_robot_task_comparison}
    \end{subfigure}
    \hfill
    \begin{subfigure}[t]{0.45\textwidth}
        \centering
        \includegraphics[width=\linewidth]{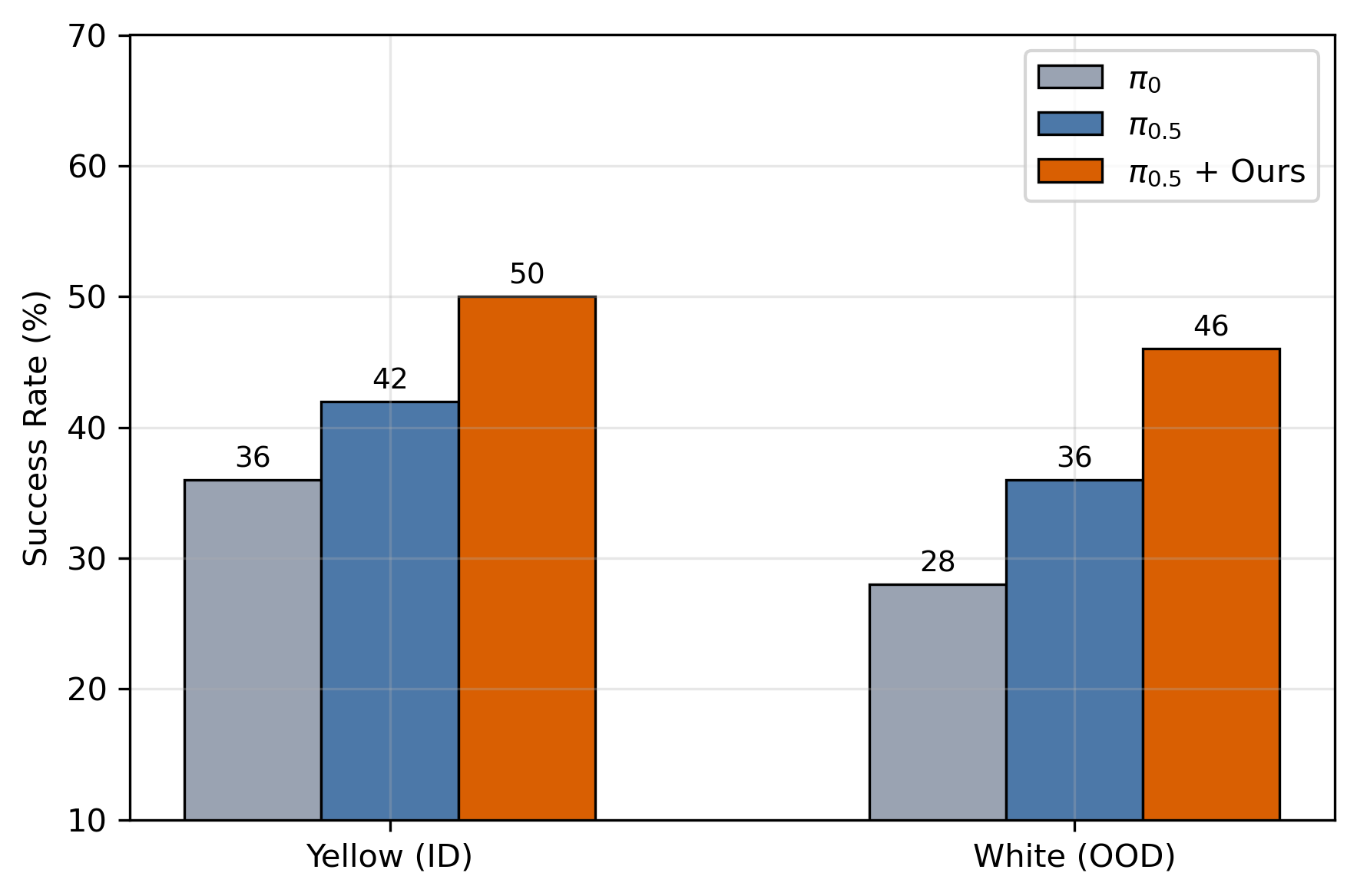}
        \caption{ALOHA-PiPER cloth folding.}
        \label{fig:aloha_piper_cloth_folding}
    \end{subfigure}
    \caption{
    Success rates on real-world robot tasks. 
    (a) OpenArm results on Bowl Stacking and Cube Handoff. 
    (b) ALOHA-PiPER results on bimanual T-shirt Folding.
    }
    \label{fig:real_robot_results}
\end{figure}

\subsection{Real-World Experiments}
\subsubsection{Setup}
We evaluate our method on two real-world bimanual platforms: an OpenArm-based dual-arm system~\citep{openarm} and an ALOHA-PiPER system~\citep{aloha,aloha2}. 
We collect 100 training trajectories per task and evaluate four tasks: bowl stacking, cube handoff, and sequential test-tube placement on OpenArm, and bimanual T-shirt folding on ALOHA-PiPER. 
The tasks cover long-horizon manipulation, bimanual coordination, fine-grained placement, deformable-object manipulation, and appearance shifts. 
Hardware details, task definitions, and training/testing protocols are provided in Appendix~\ref{app:real_robot_details}.

\subsubsection{Results}
\paragraph{OpenArm results.}
The OpenArm results are reported in Figure~\ref{fig:real_robot_results}(a) and Table~\ref{tab:real_robot_test_tube}. 
As shown in Figure~\ref{fig:real_robot_results}(a), $\pi_{0.5}$ + Ours improves the success rate from 72.0\% to 80.0\% on Bowl Stacking and from 40.0\% to 52.0\% on Cube Handoff. 
For Sequential Test-Tube Placement, Table~\ref{tab:real_robot_test_tube} shows that our method improves all completion stages, increasing the full 4/4 success rate from 18.0\% to 24.0\% and the average completed length from 1.54 to 1.94. 
These results indicate that online success memory improves execution stability in long-horizon and bimanual manipulation tasks.

\paragraph{ALOHA-PiPER results.}
The ALOHA-PiPER T-shirt folding results are shown in Figure~\ref{fig:real_robot_results}(b).
On in-domain yellow T-shirts, $\pi_{0.5}$ + Ours improves the success rate from 42.0\% to 50.0\%.
Under out-of-domain white T-shirts, our method improves the success rate from 36.0\% to 46.0\%.
Averaged across both settings, it increases performance from 39.0\% to 48.0\%, indicating improved robustness to appearance shifts in deformable-object manipulation.

\section{Ablation Studies and Analyses}
\begin{figure}[t]
    \centering
    \begin{subfigure}[t]{0.50\textwidth}
        \centering
        \includegraphics[width=\linewidth]{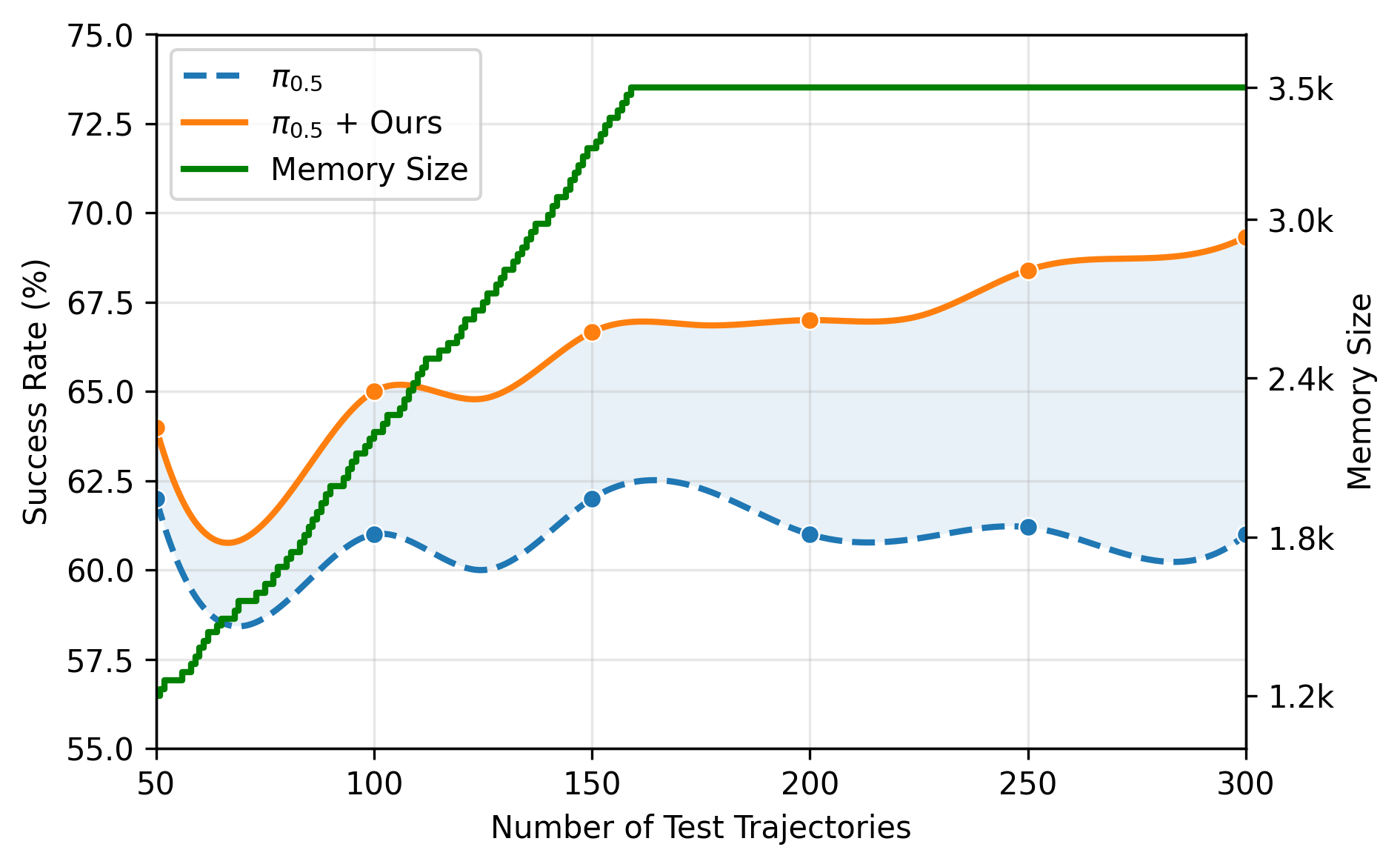}
        \caption{Cumulative success curve.}
        \label{fig:cumulative_success_curve}
    \end{subfigure}
    \hfill
    \begin{subfigure}[t]{0.47\textwidth}
        \centering
        \includegraphics[width=\linewidth]{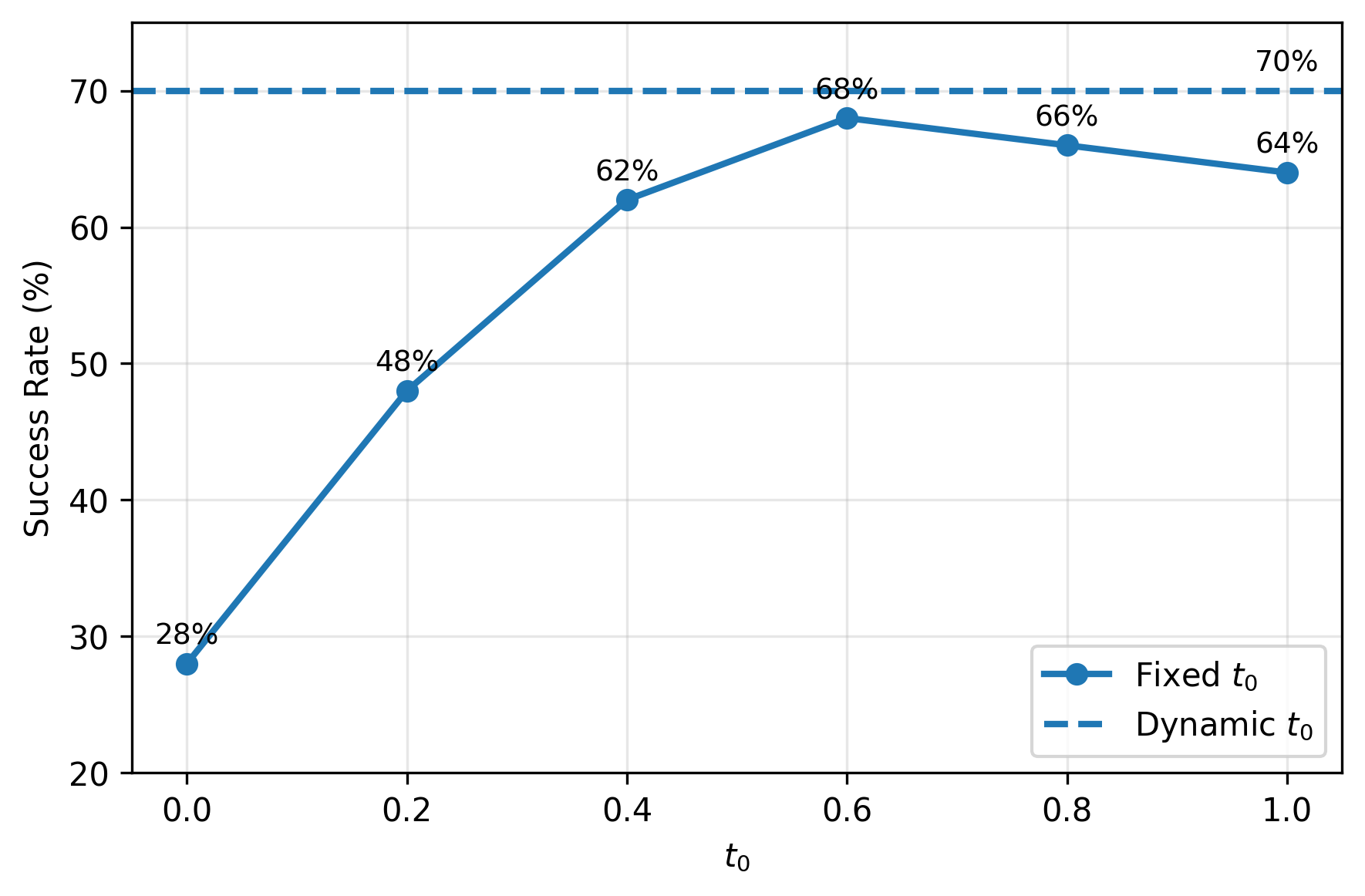}
        \caption{Effect of $t_0$.}
        \label{fig:t0_ablation}
    \end{subfigure}
    \caption{Analysis of continuous deployment and confidence-adaptive prior guidance.}
    \label{fig:deployment_and_t0}
\end{figure}

\paragraph{Continuous deployment analysis.}
\paragraph{Continuous deployment analysis.}
We evaluate continuous deployment on the \textit{Moka Pots on Stove} task from LIBERO-10. 
Starting from an empty memory, the policy is tested for 300 trajectories across different random seeds. 
During this process, verified successful observation--action segments are progressively written into the online memory and retrieved in later episodes to guide the frozen VLA. 
We use a bounded memory of 3.5k entries with FIFO replacement once the capacity is reached. 
As shown in Figure~\ref{fig:deployment_and_t0}(a), the cumulative success rate of the original $\pi_{0.5}$ remains stable, while $\pi_{0.5}$ + Ours gradually improves as the memory accumulates reusable experience and then stabilizes at a higher level. 
Notably, the gain is maintained after the memory saturates, showing that our method can exploit recent and relevant successful experience under a finite memory budget. 
A detailed memory-capacity ablation is provided in Appendix~\ref{app:Effect of Memory Capacity}.

\paragraph{Component ablation.}
We conduct component ablations on LIBERO-10 using $\pi_{0.5}$ as the frozen base policy. 
As shown in Table~\ref{tab:component_ablation}, Top-1 Retrieval improves the average success rate from 92.4\% to 93.6\%, showing that a retrieved successful chunk already provides useful test-time guidance. 
Top-$K$ Soft Aggregation further increases the success rate to 94.0\%, and the full method reaches 94.4\%, indicating that multi-candidate aggregation and DTW-based filtering improve the retrieved prior.
We also compare different prior-usage strategies with the same Top-$K$ + DTW prior. 
Direct Replay drops performance to 87.8\%, suggesting that retrieved actions should not be directly executed. 
Output Interpolation improves over the base policy to 93.0\%, but remains below intermediate initialization. 
These results show that injecting the prior into the generative sampler is more effective than post-hoc action reuse, as it allows the frozen VLA to refine actions under the current observation.
~




\begin{table}[t]
\centering
\caption{Ablation of success-memory construction on LIBERO-10 and evaluation of the success discriminator.}
\label{tab:ablation_memory_construction}
\resizebox{0.9\columnwidth}{!}{
\begin{tabular}{llcc}
\toprule
\multicolumn{4}{c}{\textbf{Success-Memory Construction}} \\
\midrule
\textbf{Memory Type}
& \textbf{Label Source}
& \textbf{Progress-Peak Truncation}
& \textbf{Avg. Success} \\
\midrule
No Memory
& --
& --
& 92.4 \\

Unverified Memory
& None
& $\times$
& 87.6 \\

Oracle Successful Full Trajectory
& Environment Label
& $\checkmark$
& \textbf{94.8} \\

Predicted Successful Full Trajectory
& Discriminator
& $\times$
& 91.8 \\

Predicted Successful Prefix Memory
& Discriminator
& $\checkmark$
& 94.4 \\
\midrule
\multicolumn{4}{c}{\textbf{Success Discriminator Evaluation}} \\
\midrule
\textbf{Metric}
& \textbf{Value}
& \textbf{Metric}
& \textbf{Value} \\
\midrule
Accuracy & 0.676 & Precision & 0.970 \\
Recall & 0.678 & F1-score & 0.798 \\
\bottomrule
\end{tabular}
}
\end{table}

\paragraph{Effect of $t_0$.}
We analyze the effect of $t_0$ on prior-guidance strength using the Moka Pots on Stove task. Fixed $t_0$ shows clear sensitivity: a large $t_0$ underuses the retrieved prior, while a small $t_0$ over-constrains generation and weakens observation-conditioned refinement. The best fixed setting achieves a success rate of 0.68 at $t_0=0.6$, whereas dynamic $t_0$ further improves it to 0.70, showing the advantage of confidence-adaptive guidance over manual tuning.

\paragraph{Effect of success-memory construction.}
Table~\ref{tab:ablation_memory_construction} shows that memory quality is critical. Storing all trajectories without verification reduces the average success rate from 92.4\% to 87.6\%, indicating that noisy test-time experience can introduce harmful priors. Using predicted successful full trajectories also underperforms the base policy, suggesting that untrimmed trajectories may contain redundant or regressive segments. In contrast, our predicted successful prefix memory achieves 94.4\%, close to the oracle memory result of 94.8\%. These results confirm the importance of both reliable success verification and progress-peak truncation. Although the discriminator has moderate recall, its high precision of 0.970 is more desirable for memory construction, since a smaller but cleaner memory is preferable to one contaminated by false successful segments.

\section{Conclusion}
We propose an online success-memory guided test-time adaptation method for generative VLAs. During continuous deployment, successful observation--action segments are stored and retrieved as action priors to initialize the generative sampler. This enables a frozen VLA to reuse environment-specific experience without parameter updates. Experiments in simulation and real-world bimanual manipulation show improved stability and success rates, especially on long-horizon tasks.

{\small
\bibliographystyle{plainnat}
\bibliography{references}

@article{rt-1,
  title={Rt-1: Robotics transformer for real-world control at scale},
  author={Brohan, Anthony and Brown, Noah and Carbajal, Justice and Chebotar, Yevgen and Dabis, Joseph and Finn, Chelsea and Gopalakrishnan, Keerthana and Hausman, Karol and Herzog, Alex and Hsu, Jasmine and others},
  journal={arXiv preprint arXiv:2212.06817},
  year={2022}
}

@inproceedings{rt-2,
  title={Rt-2: Vision-language-action models transfer web knowledge to robotic control},
  author={Zitkovich, Brianna and Yu, Tianhe and Xu, Sichun and Xu, Peng and Xiao, Ted and Xia, Fei and Wu, Jialin and Wohlhart, Paul and Welker, Stefan and Wahid, Ayzaan and others},
  booktitle={Conference on Robot Learning},
  pages={2165--2183},
  year={2023},
  organization={PMLR}
}

@article{openvla,
  title={Openvla: An open-source vision-language-action model},
  author={Kim, Moo Jin and Pertsch, Karl and Karamcheti, Siddharth and Xiao, Ted and Balakrishna, Ashwin and Nair, Suraj and Rafailov, Rafael and Foster, Ethan and Lam, Grace and Sanketi, Pannag and others},
  journal={arXiv preprint arXiv:2406.09246},
  year={2024}
}

@article{pi_0,
  title={{$\pi_0$}: A Vision-Language-Action Flow Model for General Robot Control},
  author={Black, Kevin and Brown, Noah and Driess, Danny and Esmail, Adnan and Equi, Michael and Finn, Chelsea and Fusai, Niccolo and Groom, Lachy and Hausman, Karol and Ichter, Brian and others},
  journal={arXiv preprint arXiv:2410.24164},
  year={2024}
}

@article{diffusion,
  title={Diffusion policy: Visuomotor policy learning via action diffusion},
  author={Chi, Cheng and Xu, Zhenjia and Feng, Siyuan and Cousineau, Eric and Du, Yilun and Burchfiel, Benjamin and Tedrake, Russ and Song, Shuran},
  journal={The International Journal of Robotics Research},
  volume={44},
  number={10-11},
  pages={1684--1704},
  year={2025},
  publisher={Sage Publications Sage UK: London, England}
}

@article{pi05,
  title={{$\pi_{0.5}$}: a Vision-Language-Action Model with Open-World Generalization},
  author={Intelligence, Physical and Black, Kevin and Brown, Noah and Darpinian, James and Dhabalia, Karan and Driess, Danny and Esmail, Adnan and Equi, Michael and Finn, Chelsea and Fusai, Niccolo and others},
  journal={arXiv preprint arXiv:2504.16054},
  year={2025}
}

@article{cogact,
  title={Cogact: A foundational vision-language-action model for synergizing cognition and action in robotic manipulation},
  author={Li, Qixiu and Liang, Yaobo and Wang, Zeyu and Luo, Lin and Chen, Xi and Liao, Mozheng and Wei, Fangyun and Deng, Yu and Xu, Sicheng and Zhang, Yizhong and others},
  journal={arXiv preprint arXiv:2411.19650},
  year={2024}
}

@inproceedings{dita,
  title={Dita: Scaling diffusion transformer for generalist vision-language-action policy},
  author={Hou, Zhi and Zhang, Tianyi and Xiong, Yuwen and Duan, Haonan and Pu, Hengjun and Tong, Ronglei and Zhao, Chengyang and Zhu, Xizhou and Qiao, Yu and Dai, Jifeng and others},
  booktitle={Proceedings of the IEEE/CVF International Conference on Computer Vision},
  pages={7686--7697},
  year={2025}
}

@article{octo,
  title={Octo: An open-source generalist robot policy},
  author={Team, Octo Model and Ghosh, Dibya and Walke, Homer and Pertsch, Karl and Black, Kevin and Mees, Oier and Dasari, Sudeep and Hejna, Joey and Kreiman, Tobias and Xu, Charles and others},
  journal={arXiv preprint arXiv:2405.12213},
  year={2024}
}

@article{pi0fast,
  title={Fast: Efficient action tokenization for vision-language-action models},
  author={Pertsch, Karl and Stachowicz, Kyle and Ichter, Brian and Driess, Danny and Nair, Suraj and Vuong, Quan and Mees, Oier and Finn, Chelsea and Levine, Sergey},
  journal={arXiv preprint arXiv:2501.09747},
  year={2025}
}

@article{Libero-plus,
  title={Libero-plus: In-depth robustness analysis of vision-language-action models},
  author={Fei, Senyu and Wang, Siyin and Shi, Junhao and Dai, Zihao and Cai, Jikun and Qian, Pengfang and Ji, Li and He, Xinzhe and Zhang, Shiduo and Fei, Zhaoye and others},
  journal={arXiv preprint arXiv:2510.13626},
  year={2025}
}

@article{romer2025failure,
  title={Failure prediction at runtime for generative robot policies},
  author={R{\"o}mer, Ralf and Kobras, Adrian and Worbis, Luca and Schoellig, Angela P},
  journal={arXiv preprint arXiv:2510.09459},
  year={2025}
}

@article{libero,
  title={Libero: Benchmarking knowledge transfer for lifelong robot learning},
  author={Liu, Bo and Zhu, Yifeng and Gao, Chongkai and Feng, Yihao and Liu, Qiang and Zhu, Yuke and Stone, Peter},
  journal={Advances in Neural Information Processing Systems},
  volume={36},
  pages={44776--44791},
  year={2023}
}

@article{simpler,
  title={Evaluating real-world robot manipulation policies in simulation},
  author={Li, Xuanlin and Hsu, Kyle and Gu, Jiayuan and Pertsch, Karl and Mees, Oier and Walke, Homer Rich and Fu, Chuyuan and Lunawat, Ishikaa and Sieh, Isabel and Kirmani, Sean and others},
  journal={arXiv preprint arXiv:2405.05941},
  year={2024}
}

@inproceedings{Open-x-embodiment,
  title={Open x-embodiment: Robotic learning datasets and rt-x models: Open x-embodiment collaboration 0},
  author={O’Neill, Abby and Rehman, Abdul and Maddukuri, Abhiram and Gupta, Abhishek and Padalkar, Abhishek and Lee, Abraham and Pooley, Acorn and Gupta, Agrim and Mandlekar, Ajay and Jain, Ajinkya and others},
  booktitle={2024 IEEE International Conference on Robotics and Automation (ICRA)},
  pages={6892--6903},
  year={2024},
  organization={IEEE}
}

@inproceedings{reduction_rl,
  title={A reduction of imitation learning and structured prediction to no-regret online learning},
  author={Ross, St{\'e}phane and Gordon, Geoffrey and Bagnell, Drew},
  booktitle={Proceedings of the fourteenth international conference on artificial intelligence and statistics},
  pages={627--635},
  year={2011},
  organization={JMLR Workshop and Conference Proceedings}
}

@inproceedings{Hg-dagger,
  title={Hg-dagger: Interactive imitation learning with human experts},
  author={Kelly, Michael and Sidrane, Chelsea and Driggs-Campbell, Katherine and Kochenderfer, Mykel J},
  booktitle={2019 International Conference on Robotics and Automation (ICRA)},
  pages={8077--8083},
  year={2019},
  organization={IEEE}
}

@article{robomonkey,
  title={Robomonkey: Scaling test-time sampling and verification for vision-language-action models},
  author={Kwok, Jacky and Agia, Christopher and Sinha, Rohan and Foutter, Matt and Li, Shulu and Stoica, Ion and Mirhoseini, Azalia and Pavone, Marco},
  journal={arXiv preprint arXiv:2506.17811},
  year={2025}
}

@article{Verifier-free,
  title={Verifier-free Test-Time Sampling for Vision Language Action Models},
  author={Jang, Suhyeok and Kim, Dongyoung and Kim, Changyeon and Kim, Youngsuk and Shin, Jinwoo},
  journal={arXiv preprint arXiv:2510.05681},
  year={2025}
}

@article{rover,
  title={RoVer: Robot Reward Model as Test-Time Verifier for Vision-Language-Action Model},
  author={Dai, Mingtong and Liu, Lingbo and Bai, Yongjie and Liu, Yang and Wang, Zhouxia and Su, Rui and Chen, Chunjie and Lin, Liang and Wu, Xinyu},
  journal={arXiv preprint arXiv:2510.10975},
  year={2025}
}

@article{taco,
  title={Steering Vision-Language-Action Models as Anti-Exploration: A Test-Time Scaling Approach},
  author={Yang, Siyuan and Zhang, Yang and He, Haoran and Pan, Ling and Li, Xiu and Bai, Chenjia and Li, Xuelong},
  journal={arXiv preprint arXiv:2512.02834},
  year={2025}
}

@article{strap,
  title={Strap: Robot sub-trajectory retrieval for augmented policy learning},
  author={Memmel, Marius and Berg, Jacob and Chen, Bingqing and Gupta, Abhishek and Francis, Jonathan},
  journal={arXiv preprint arXiv:2412.15182},
  year={2024}
}

@article{memoryvla,
  title={Memoryvla: Perceptual-cognitive memory in vision-language-action models for robotic manipulation},
  author={Shi, Hao and Xie, Bin and Liu, Yingfei and Sun, Lin and Liu, Fengrong and Wang, Tiancai and Zhou, Erjin and Fan, Haoqiang and Zhang, Xiangyu and Huang, Gao},
  journal={arXiv preprint arXiv:2508.19236},
  year={2025}
}

@article{retrieval-skill-base,
  title={Learning and retrieval from prior data for skill-based imitation learning},
  author={Nasiriany, Soroush and Gao, Tian and Mandlekar, Ajay and Zhu, Yuke},
  journal={arXiv preprint arXiv:2210.11435},
  year={2022}
}

@inproceedings{retrieval,
  title={Retrieval-augmented embodied agents},
  author={Zhu, Yichen and Ou, Zhicai and Mou, Xiaofeng and Tang, Jian},
  booktitle={Proceedings of the IEEE/CVF Conference on Computer Vision and Pattern Recognition},
  pages={17985--17995},
  year={2024}
}

@article{conrft,
  title={Conrft: A reinforced fine-tuning method for vla models via consistency policy},
  author={Chen, Yuhui and Tian, Shuai and Liu, Shugao and Zhou, Yingting and Li, Haoran and Zhao, Dongbin},
  journal={arXiv preprint arXiv:2502.05450},
  year={2025}
}

@article{expres-vla,
  title={ExpReS-VLA: Specializing Vision-Language-Action Models Through Experience Replay and Retrieval},
  author={Syed, Shahram Najam and Ahuja, Yatharth and Jakobsson, Arthur and Ichnowski, Jeff},
  journal={arXiv preprint arXiv:2511.06202},
  year={2025}
}

@article{continually,
  title={Continually Evolving Skill Knowledge in Vision Language Action Model},
  author={Wu, Yuxuan and Wang, Guangming and Yang, Zhiheng and Yao, Maoqing and Sheil, Brian and Wang, Hesheng},
  journal={arXiv preprint arXiv:2511.18085},
  year={2025}
}

@article{aloha,
  title={Learning fine-grained bimanual manipulation with low-cost hardware},
  author={Zhao, Tony Z and Kumar, Vikash and Levine, Sergey and Finn, Chelsea},
  journal={arXiv preprint arXiv:2304.13705},
  year={2023}
}

@article{aloha2,
  title={Aloha 2: An enhanced low-cost hardware for bimanual teleoperation},
  author={Zhao, TZ and Schmidgall, S and Kim, JW and Deguet, A and Kobilarov, M and Krieger, A and Finn, C},
  journal={arXiv preprint arXiv:2405.02292},
  year={2024}
}

@misc{openarm,
  title        = {OpenArm: A Fully Open-Source Humanoid Robot Arm for Physical AI Research},
  author       = {{Enactic, Inc.}},
  year         = {2025},
  howpublished = {\url{https://openarm.dev/}},
  note         = {Accessed: 2026-05-05}
}

@article{vlac,
  title={A vision-language-action-critic model for robotic real-world reinforcement learning},
  author={Zhai, Shaopeng and Zhang, Qi and Zhang, Tianyi and Huang, Fuxian and Zhang, Haoran and Zhou, Ming and Zhang, Shengzhe and Liu, Litao and Lin, Sixu and Pang, Jiangmiao},
  journal={arXiv preprint arXiv:2509.15937},
  year={2025}
}
}

\newpage
\appendix
\section{Details of the Progress Estimator}
\label{app:progress_estimator}

\paragraph{Model overview.}
We use a pretrained VLAC critic~\citep{vlac} as the progress estimator for automatic trajectory evaluation and online success-memory construction.
VLAC is a vision-language-action-critic model built upon InternVL, where action generation and process evaluation are unified in a multimodal autoregressive framework.
In this work, we only use its critic capability.
Given two visual observations and a language instruction, the critic predicts a signed progress change indicating whether the second observation is closer to the task goal than the first one.
A positive value indicates forward task progress, a negative value indicates regression or deviation, and a value close to zero indicates little task-relevant change.
Compared with single-state success classification, this pair-wise progress formulation provides a more fine-grained signal for long-horizon manipulation.

VLAC critic is trained from temporal supervision in successful task videos.
Given a trajectory $O=(o_1,\ldots,o_T)$ with task instruction $l_{\mathrm{task}}$, two frames $o_i$ and $o_{i+\Delta t}$ are sampled, and the progress label is constructed from their temporal offset:
\begin{equation}
c_{i,i+\Delta t}
=
\frac{\Delta t}{T-i}.
\end{equation}
Forward pairs correspond to positive progress, while reversed pairs provide negative progress samples.
The training further includes static-frame filtering, forward/backward joint sampling, task-completion prediction, and semantically mismatched samples, which improve robustness to stagnation, regression, and task-irrelevant visual changes.
Since this learning process mainly depends on visual states and language goals rather than a unified action space, VLAC can be trained with heterogeneous human and robot trajectory data and generalizes across different embodiments and scenes.

\paragraph{Role of the reference video.}
VLAC supports in-context progress understanding through a reference process.
For unseen tasks, scenes, or embodiments, a language instruction alone may not fully specify the task stages and completion condition.
A reference process provides an execution example that reveals the expected temporal structure, key visual transitions, and task logic.
Formally, VLAC estimates progress with an additional reference sequence as
\begin{equation}
c_{i,i+\Delta t}
=
\mathrm{VLAC}
\left(
o_i,o_{i+\Delta t};
l_{\mathrm{task}}, O_{\mathrm{ref}}, o_0
\right),
\end{equation}
where $O_{\mathrm{ref}}$ can be a robot or human demonstration, and $o_0$ denotes the initial observation of the current trajectory.
The reference process helps align the current execution with a successful example and enables one-shot transfer to new task instances.

In our framework, we provide one successful demonstration video for each task as the reference process $R$.
The reference video is not used as action supervision and does not update the policy.
It only serves as an in-context example for the VLAC critic.
For example, in pick-and-place tasks, it implicitly specifies the stage order such as approaching, grasping, moving, and placing.
This helps the critic distinguish forward progress from stagnation, regression, and failure more reliably than using the language instruction alone.

\paragraph{Usage in our framework.}
For the $i$-th test trajectory, we evaluate progress at a fixed interval $\Delta$.
Let $\tau_m$ and $\tau_{m-1}$ denote two adjacent evaluation timesteps.
Conditioned on the instruction $l^{(i)}$ and the reference process $R$, the interval-level progress is computed as
\begin{equation}
c^{(i)}_{\tau_m}
=
\Phi_\psi
\left(
o^{(i)}_{\tau_{m-1}},
o^{(i)}_{\tau_m},
l^{(i)};R
\right).
\end{equation}
These interval-level scores are accumulated into a trajectory-level progress score.
Instead of using the terminal progress, we take the maximum accumulated progress as the completion score, which is more robust to overshooting, collisions, or redundant motions after near completion.
If the maximum progress exceeds a threshold, the trajectory is considered to contain reusable successful experience.
We then store only the observation-action prefix before the progress peak into the online memory, filtering out failure segments, regressions, and post-success redundancy.

\paragraph{Why VLAC critic.}
VLAC critic is well suited to our setting because our goal is not episode-level success labeling, but conservative extraction of reusable successful segments.
The quality of online memory directly affects the reliability of the retrieved action prior.
Thus, the estimator should identify task-relevant progress while avoiding the inclusion of failed or regressive segments.
VLAC provides such a task-conditioned pair-wise progress signal and can use a reference video as an in-context task prior without training a task-specific classifier.

This design is also compatible with frozen-VLA test-time adaptation.
The progress estimator is pretrained and external to the base policy, requiring no policy parameter update and no assumption about the action representation of the underlying generative VLA.
Our discriminator analysis further supports this choice.
Although the overall accuracy and recall are moderate, the precision reaches $0.970$.
This means that the estimator may miss some successful trajectories, but trajectories predicted as successful are highly reliable.
For success-memory construction, high precision is more important than high recall: missing a successful segment only slows memory growth, whereas storing a failed segment can contaminate memory and induce misleading action priors.
Empirically, storing all trajectories without verification reduces the average success rate from $92.4\%$ to $87.6\%$, while the predicted successful prefix memory achieves $94.4\%$, close to the oracle memory result of $94.8\%$.
This indicates that a smaller but cleaner memory is preferable to a larger noisy memory for retrieve-then-steer test-time adaptation.

\section{Prior Guidance for Diffusion Action Heads}
\label{app:diffusion_prior_guidance}
The main paper presents confidence-adaptive prior guidance using a
flow-matching action head. For VLA policies equipped with diffusion-based action
heads, the retrieved elite action prior $a_{\mathrm{elite}}$ can also be used to
guide action generation. The key idea is to first perturb
$a_{\mathrm{elite}}$ to an intermediate noise level following the diffusion
forward process, and then let the original diffusion action generator perform
the remaining conditional denoising steps.

\subsection{Diffusion Action Generation}

For diffusion-based action heads, an action chunk is modeled as a random
variable generated by gradually denoising Gaussian noise. Given the conditioning
feature $z_t$ corresponding to the current observation and task instruction,
standard diffusion sampling starts from
\begin{equation}
x_N \sim \mathcal{N}(0,I),
\end{equation}
and iteratively applies the reverse denoising process:
\begin{equation}
x_{n-1}
\sim
p_\theta(x_{n-1}\mid x_n,z_t),
\qquad
n=N,N-1,\ldots,1.
\end{equation}
The final action chunk is obtained as
\begin{equation}
\hat{a}_{t:t+H-1}
=
x_0 .
\end{equation}
Here, $N$ denotes the number of diffusion sampling steps, $H$ denotes the action
prediction horizon, and $z_t$ is the conditional feature extracted by the VLA
from the current observation $o_t$ and language instruction $l$. Since standard
diffusion sampling starts entirely from random noise, the generated action may
be sensitive to noise initialization and local state deviations. To exploit
environment-specific experience stored in the online success memory, we use the
retrieved elite prior $a_{\mathrm{elite}}$ to guide the diffusion denoising
process.

\subsection{Prior Perturbation and Conditional Denoising}

For DDPM/DDIM-style diffusion action heads, the forward noising process can be
written as
\begin{equation}
q(x_n\mid x_0)
=
\mathcal{N}
\left(
x_n;
\sqrt{\bar{\alpha}_n}x_0,
(1-\bar{\alpha}_n)I
\right),
\end{equation}
or equivalently,
\begin{equation}
x_n
=
\sqrt{\bar{\alpha}_n}x_0
+
\sqrt{1-\bar{\alpha}_n}\epsilon,
\qquad
\epsilon\sim\mathcal{N}(0,I).
\end{equation}
Here, $\bar{\alpha}_n$ denotes the cumulative signal coefficient in the
diffusion noise schedule. As $n$ increases, $\bar{\alpha}_n$ decreases and the
sample moves from an action-like state toward a high-noise state.

When the retrieval module produces an elite action prior
$a_{\mathrm{elite}}$, we regard it as a local successful action endpoint and
perturb it to an intermediate diffusion step $n_0$ using the original forward
noising process:
\begin{equation}
x_{n_0}
=
\sqrt{\bar{\alpha}_{n_0}}a_{\mathrm{elite}}
+
\sqrt{1-\bar{\alpha}_{n_0}}\epsilon,
\qquad
\epsilon\sim\mathcal{N}(0,I).
\end{equation}
The intermediate step $n_0$ is determined by the confidence-adaptive guidance
schedule introduced in the main text. Intuitively, a more reliable retrieved
prior corresponds to a lower noise level and thus stronger prior guidance,
whereas a less reliable prior corresponds to a higher noise level and leaves
more freedom for the diffusion model.

The sampler then starts from $x_{n_0}$ instead of $x_N$ and performs the
remaining reverse denoising steps:
\begin{equation}
x_{n-1}
\sim
p_\theta(x_{n-1}\mid x_n,z_t),
\qquad
n=n_0,n_0-1,\ldots,1.
\end{equation}
The final generated action chunk is
\begin{equation}
\hat{a}_{t:t+H-1}
=
x_0 .
\end{equation}

This procedure provides a diffusion-compatible form of prior guidance:
$a_{\mathrm{elite}}$ is first mapped to an intermediate noisy state on the
diffusion trajectory, and the original diffusion policy then completes
conditional denoising under the current observation feature $z_t$. Therefore,
the model does not simply replay the retrieved action; instead, it generates an
action chunk near the neighborhood of historical successful behavior while
remaining conditioned on the current state. If the online success memory is
empty or the retrieved candidates fail the similarity and trajectory-consistency
filters, the system does not construct $a_{\mathrm{elite}}$ and falls back to
standard diffusion sampling from $x_N\sim\mathcal{N}(0,I)$.

\section{Component-Aware Aggregation of Action Priors}
\label{app:component_aggregation}

\paragraph{Action component decomposition.}
Since the action spaces of different robotic platforms may contain different
types of control variables, directly applying element-wise linear averaging to
the entire action vector is not always appropriate. We therefore adopt a
component-aware aggregation strategy for constructing the elite action prior.
For the $h$-th prediction step of the $i$-th candidate action chunk, we write
\begin{equation}
a_{i,h}
=
(\Delta p_{i,h}, \Delta r_{i,h}, g_{i,h}),
\end{equation}
where $\Delta p_{i,h}$ denotes action components in Euclidean spaces, such as
end-effector position increments or joint-angle increments; $\Delta r_{i,h}$
denotes orientation or orientation-increment components; and $g_{i,h}$ denotes
the gripper command.

\paragraph{Aggregation of Euclidean action components.}
For action components that lie in Euclidean spaces, including end-effector
positions, joint angles, and their corresponding increments, we directly apply
similarity-weighted averaging:
\begin{equation}
\Delta p_{\mathrm{elite},h}
=
\sum_{i\in\mathcal{I}} w_i \Delta p_{i,h}.
\end{equation}
This operation is suitable for action dimensions with a linear structure and
forms a smooth local action prior from multiple similar successful behaviors.

\paragraph{Aggregation of orientation components.}
For orientations and orientation increments, direct linear averaging may violate
the geometry of the rotation space. Therefore, we first map each axis-angle
orientation increment to the rotation group $SO(3)$:
\begin{equation}
R_{i,h}
=
\mathrm{Exp}(\Delta r_{i,h}),
\end{equation}
where $\mathrm{Exp}(\cdot)$ denotes the exponential map from the Lie algebra
$\mathfrak{so}(3)$ to $SO(3)$. We then compute the weighted geodesic mean on
$SO(3)$:
\begin{equation}
R_{\mathrm{elite},h}
=
\arg\min_{R\in SO(3)}
\sum_{i\in\mathcal{I}}
w_i
\left\|
\mathrm{Log}(R_{i,h}^{\top}R)
\right\|_2^2,
\end{equation}
where $\mathrm{Log}(\cdot)$ denotes the logarithm map from $SO(3)$ to
$\mathfrak{so}(3)$. Finally, the averaged rotation is mapped back to the
axis-angle representation:
\begin{equation}
\Delta r_{\mathrm{elite},h}
=
\mathrm{Log}(R_{\mathrm{elite},h}).
\end{equation}
This procedure ensures that orientation aggregation respects the geometry of
the rotation manifold and avoids the inconsistency that may arise from naive
linear averaging.

\paragraph{Aggregation of gripper actions.}
Since different robotic platforms parameterize gripper commands differently, we
use an adaptive aggregation rule for gripper actions. For continuous gripper
commands, such as gripper width, finger-joint position, or normalized continuous
control values, we apply weighted averaging:
\begin{equation}
g_{\mathrm{elite},h}
=
\sum_{i\in\mathcal{I}} w_i g_{i,h},
\end{equation}
and clip the result to the valid action range. For discrete gripper commands,
such as open-or-close commands, we use weighted voting:
\begin{equation}
P_h(c)
=
\sum_{i\in\mathcal{I}}
w_i
\mathbb{I}[g_{i,h}=c],
\end{equation}
\begin{equation}
g_{\mathrm{elite},h}
=
\arg\max_c P_h(c).
\end{equation}
When the retrieved gripper commands exhibit strong conflicts between opening and
closing directions, we adopt a conservative fallback strategy: the gripper
command from the most similar nearest-neighbor candidate is used. This avoids
averaging contradictory gripper actions into an ambiguous gripper prior.

\paragraph{Final action prior.}
After component-aware aggregation, the elite action prior at the $h$-th
prediction step is written as
\begin{equation}
a_{\mathrm{elite},h}
=
(
\Delta p_{\mathrm{elite},h},
\Delta r_{\mathrm{elite},h},
g_{\mathrm{elite},h}
).
\end{equation}
By concatenating the priors over all prediction steps, we obtain the complete
elite action prior:
\begin{equation}
a_{\mathrm{elite}}
=
\{a_{\mathrm{elite},h}\}_{h=1}^{H}.
\end{equation}
This design allows our method to support end-effector pose control, joint-space
control, and different gripper parameterizations, thereby avoiding overfitting
the method formulation to a specific dataset or robotic platform.

\section{Real-World Robot Experiment Details}
\label{app:real_robot_details}

\subsection{Hardware Platforms and Camera Configuration}

We evaluate our method on two real-world bimanual robot platforms: an OpenArm-based dual-arm system and an ALOHA-PiPER system. The OpenArm platform consists of two 7-DoF humanoid robot arms equipped with parallel grippers, while the ALOHA-PiPER platform follows an ALOHA-style bimanual setup and uses two 6-DoF AgileX PiPER arms with two-finger grippers. These two platforms provide complementary testbeds for evaluating our method across different arm kinematics, gripper designs, and workspace layouts.

Both platforms take multi-view RGB images and proprioceptive states as policy inputs. Specifically, each platform is equipped with three camera views: two wrist-mounted cameras attached to the left and right grippers, and one external third-person camera. The wrist cameras provide close-up observations of local manipulation regions, such as grasping, handoff, and placement areas, while the third-person camera provides a global view of the workspace, including object layouts, arm configurations, and overall task progress.

The placement of the third-person camera differs slightly between the two platforms due to hardware constraints. On the ALOHA-PiPER platform, the external camera is mounted above the center region between the two arms on the same side as the robot arms, providing a direct overhead view of the T-shirt manipulation area. On the OpenArm platform, due to the limited mounting space on the robot side, the external camera is placed above the center region between the two arms but on the opposite side of the workspace. Although the camera placements are different, both configurations provide complementary global observations together with the two wrist-mounted views, enabling the policy to perceive both fine-grained local interactions and long-horizon task progress.

\subsection{Task Definitions}

We design four real-world manipulation tasks to evaluate the proposed method, covering long-horizon object manipulation, bimanual coordination, fine-grained placement, and deformable-object manipulation.

\paragraph{Bowl Stacking.}
In this task, the two robot arms grasp one bowl from each side of the workspace and stack the two bowls in the center region. The task requires the policy to coordinate both arms to complete grasping, transportation, alignment, and stacking. This task mainly evaluates long-horizon object manipulation, bimanual spatial coordination, and relative pose alignment between objects. The main challenges come from unstable bowl grasps, the need for accurate rim alignment, and the risk of collision or slippage during the stacking stage.

\paragraph{Cube Handoff.}
In this task, the right arm first picks up a red cube from a bowl on the right side of the workspace, transfers it to the left arm, and the left arm then places it into a bowl on the left side. This task evaluates bimanual handoff ability, including grasping, approaching, pose alignment between two grippers, object transfer, release timing, and final placement. Compared with single-arm pick-and-place tasks, Cube Handoff requires more precise coordination between the two end-effectors. A small error in relative gripper pose or release timing can easily cause the cube to drop.

\paragraph{Sequential Test-Tube Placement.}
In this task, the right arm sequentially picks up four test tubes arranged on the table from left to right and places them into a test-tube rack. This task evaluates long-horizon sequential manipulation, fine-grained grasping, and precise placement into narrow slots. Since test tubes are thin and elongated objects, they are difficult to grasp stably, and the rack holes impose strict requirements on placement position and orientation. To improve task success and better exploit the bimanual setup, we keep the left arm stationary during execution and orient its wrist camera toward the test-tube rack. This provides a fine-grained local view of the rack region, helping the policy observe the target holes more accurately. The main challenges include accumulated errors over repeated operations, unstable grasping of thin objects, precise insertion, and state drift during long-horizon execution.

\paragraph{Bimanual T-shirt Folding.}
On the ALOHA-PiPER platform, we evaluate a bimanual T-shirt folding task. The two arms need to collaboratively grasp key regions of the T-shirt and complete the folding process. Unlike rigid-object manipulation, T-shirt folding involves deformable-object dynamics, where the object shape changes continuously during grasping, dragging, and folding. This task therefore requires robust visual understanding and closed-loop action adjustment. Demonstrations are collected using a yellow T-shirt, while testing is conducted under both in-domain and out-of-domain settings. The in-domain setting uses yellow T-shirts, whereas the out-of-domain setting uses T-shirts with unseen colors. This design evaluates the robustness of the policy to visual appearance shifts in deformable-object manipulation. The main challenges include uncertain cloth deformation, self-occlusion, localization of key grasping regions, and visual distribution shifts caused by color changes.

\subsection{Training and Testing Protocol}

For each real-world task, we collect 100 demonstration trajectories to fine-tune the base VLA policy. We use the same basic training configuration across all tasks, with a batch size of 64 and a learning rate of $2.5 \times 10^{-5}$. The number of training steps is adjusted according to the task horizon and manipulation complexity, ranging from 8k to 10k steps. For relatively shorter tasks, such as Bowl Stacking and Cube Handoff, the policy is trained for around 8k steps. For longer-horizon or more fine-grained tasks, such as Sequential Test-Tube Placement and Bimanual T-shirt Folding, the number of training steps is increased toward 10k steps to better cover the full task procedure and key manipulation stages.

During evaluation, the base VLA policy is kept frozen, and no additional gradient updates or online fine-tuning are performed. Each real-world task is evaluated over 50 trials. For fair comparison, the baseline policy and our method are tested under the same set of initial states. We only enable the proposed online success-memory mechanism on top of the frozen policy. During the evaluation of each task, the online memory is initialized and continuously updated according to our success-memory construction mechanism. Specifically, the robot executes action chunks generated by the frozen VLA policy during continuous test-time deployment, and reusable observation-action segments are stored into the online success memory when successful experience is identified. In subsequent trials, the system retrieves relevant successful experience according to the current observation and uses the retrieved action prior to guide the generative action sampling process. This protocol ensures that the performance improvement comes from non-parametric test-time memory retrieval and prior-guided generation, rather than from additional policy training, parameter updates, or different initial-state distributions.
\section{Effect of Memory Capacity}
\label{app:Effect of Memory Capacity}
As shown in Table~\ref{tab:memory_capacity}, increasing the memory capacity consistently improves the final cumulative success rate, indicating that a larger memory provides more reusable successful experience for retrieval-guided action generation. The improvement is more pronounced when the capacity increases from 0 to 3k entries, while the gain gradually saturates beyond 4k. In particular, the 5k memory budget achieves a final cumulative success rate of 70.8\%, which is close to the unlimited-memory setting of 71.2\%. This suggests that our method does not rely on an ever-growing memory, but can achieve near-saturated performance with a bounded memory budget and FIFO replacement.
\begin{table}[t]
\centering
\caption{
Effect of memory capacity on continuous deployment. We evaluate $\pi_{0.5}$ + Ours on the \textit{Moka Pots on Stove} task for 300 test trajectories. $C$ denotes the maximum number of stored observation-action entries. When the memory exceeds $C$, FIFO eviction is applied.
}
\label{tab:memory_capacity}
\resizebox{0.78\linewidth}{!}{
\begin{tabular}{lccc}
\toprule
Memory Capacity $C$ & Eviction & Final Mem. Size & Final Cum. SR (\%) \\
\midrule
0 & -- & 0 & 61.0 \\
1k & FIFO & 1.0k & 64.0 \\
2k & FIFO & 2.0k & 66.2 \\
3k & FIFO & 3.0k & 68.5 \\
4k & FIFO & 4.0k & 70.2 \\
5k & FIFO & 5.0k & 70.8 \\
Unlimited & -- & All & 71.2 \\
\bottomrule
\end{tabular}
}
\end{table}

\section{Inference-Time Analysis}
\label{app:inference_time}

We further analyze the inference-time overhead introduced by the proposed retrieve-then-steer mechanism.
All results are reported as relative inference time normalized by the frozen base policy.
The runtime includes action generation for the current decision step. For our method, we report two
settings: one excluding retrieval overhead, which measures only the prior-guided generative sampling
process, and one including the full retrieval, filtering, aggregation, and sampling pipeline.

\begin{table}[t]
\centering
\caption{Relative inference time normalized by the frozen base policy. ``Ours w/o Retrieval''
measures the prior-guided sampler after the elite prior is available, while ``Ours Full'' includes
retrieval, filtering, prior aggregation, and action generation.}
\label{tab:inference_time}
\setlength{\tabcolsep}{8pt}
\renewcommand{\arraystretch}{1.08}
\begin{tabular}{lcc}
\toprule
\textbf{Method} & \textbf{Retrieval Included} & \textbf{Relative Time} \\
\midrule
Base VLA & -- & $1.00\times$ \\
Ours w/o Retrieval & No & $0.95\times$ \\
Ours Full & Yes & $1.10\times$ \\
\bottomrule
\end{tabular}
\end{table}

As shown in Table~\ref{tab:inference_time}, the proposed method does not introduce heavy
inference overhead. When the retrieval cost is excluded, our prior-guided sampler is slightly faster
than the base policy, reducing the normalized inference time from $1.00\times$ to $0.95\times$.
This is because the retrieved elite prior initializes the generative process from an intermediate state,
allowing the sampler to perform fewer effective generation steps than the original noise-starting
sampler.

When retrieval, trajectory-consistency filtering, and prior aggregation are included, the full method
requires $1.10\times$ the inference time of the base policy. This moderate overhead mainly comes
from nearest-neighbor search and candidate filtering in the online success memory. Since these
operations are non-parametric and do not require additional forward passes through the VLA or
parameter updates, the overall runtime remains lightweight. Combined with the success-rate gains
reported in the main experiments, these results suggest that the proposed retrieve-then-steer mechanism
offers a favorable trade-off between deployment efficiency and closed-loop reliability.

\section{Hyperparameter Sensitivity}
\label{app:hyperparameter_sensitivity}

We analyze the sensitivity of the proposed retrieve-then-steer framework to key
hyperparameters in the retrieval, memory construction, and prior-construction pipeline.
Unless otherwise specified, experiments are conducted on LIBERO-10 using $\pi_{0.5}$ as
the frozen base policy. We vary one hyperparameter at a time while keeping the others
fixed to the default setting: $K=10$, $\tau=0.05$, $\gamma_{\rm sim}=0.9992$, and
success threshold $\eta=0.95$. The base policy achieves an average success rate of
$92.4\%$.

\begin{table}[t]
\centering
\caption{Hyperparameter sensitivity on LIBERO-10. We report the average success rate (\%)
over all tasks. The default setting is highlighted in gray.}
\label{tab:hyperparameter_sensitivity}
\setlength{\tabcolsep}{6pt}
\renewcommand{\arraystretch}{1.08}
\footnotesize
\begin{tabular}{llc}
\toprule
\textbf{Hyperparameter} & \textbf{Value} & \textbf{Avg. Success (\%)} \\
\midrule
Base $\pi_{0.5}$ & -- & 92.4 \\
\midrule
\multirow{5}{*}{Top-$K$ retrieval size}
& $K=1$  & 93.6 \\
& $K=3$  & 93.8 \\
& $K=5$  & 94.1 \\
& \cellcolor{gray!15}{$K=10$} & \cellcolor{gray!15}{94.4} \\
& $K=20$ & 94.2 \\
\midrule
\multirow{5}{*}{Similarity threshold $\gamma_{\rm sim}$}
& $0.9988$ & 93.7 \\
& $0.9990$ & 94.1 \\
& \cellcolor{gray!15}{$0.9992$} & \cellcolor{gray!15}{94.4} \\
& $0.9995$ & 94.0 \\
& $0.9998$ & 93.5 \\
\midrule
\multirow{5}{*}{Aggregation temperature $\tau$}
& $0.01$ & 93.8 \\
& $0.03$ & 94.2 \\
& \cellcolor{gray!15}{$0.05$} & \cellcolor{gray!15}{94.4} \\
& $0.10$ & 94.1 \\
& $0.20$ & 93.9 \\
\midrule
\multirow{5}{*}{Success threshold $\eta$}
& $0.85$ & 93.8 \\
& $0.90$ & 94.1 \\
& \cellcolor{gray!15}{$0.95$} & \cellcolor{gray!15}{94.4} \\
& $0.975$ & 94.2 \\
& $0.99$ & 93.9 \\
\bottomrule
\end{tabular}
\end{table}

\begin{table}[t]
\centering
\caption{Default hyperparameters for adaptive prior guidance. These parameters are kept fixed
across experiments and are empirically stable within reasonable ranges.}
\label{tab:adaptive_t0_params}
\setlength{\tabcolsep}{8pt}
\renewcommand{\arraystretch}{1.08}
\footnotesize
\begin{tabular}{lc}
\toprule
\textbf{Hyperparameter} & \textbf{Value} \\
\midrule
Reference similarity $s_{\rm ref}$ & $\gamma_{\rm sim}$ \\
Similarity scale $s_{\rm scale}$ & $5\times 10^{-4}$ \\
Clipping range $c_{\max}$ & $50$ \\
Similarity coefficient $\alpha$ & $1.0$ \\
DTW-dispersion coefficient $\beta$ & $0.05$ \\
Mapping sharpness $\gamma$ & $2.0$ \\
\bottomrule
\end{tabular}
\end{table}

As shown in Table~\ref{tab:hyperparameter_sensitivity}, the proposed method is relatively
robust to a broad range of hyperparameter choices. First, increasing the retrieval size from
$K=1$ to $K=10$ consistently improves performance, indicating that aggregating multiple
successful chunks provides a more reliable action prior than using a single nearest neighbor.
However, further increasing $K$ to $20$ slightly reduces the success rate, likely because less
relevant trajectories are introduced into the candidate set.

Second, the similarity threshold $\gamma_{\rm sim}$ controls the trade-off between prior
coverage and prior quality. A lower threshold accepts more retrieved candidates but may include
mismatched action chunks, while an overly strict threshold rejects useful candidates and causes
the method to fall back to the base sampler more frequently. The default value
$\gamma_{\rm sim}=0.9992$ achieves the best balance between filtering unreliable retrievals
and preserving sufficient reusable experience.

Third, the aggregation temperature $\tau$ affects the sharpness of the similarity-based soft
weights. A very small temperature makes the aggregation close to nearest-neighbor selection,
whereas a large temperature assigns nearly uniform weights to retrieved candidates. The best
performance is obtained at $\tau=0.05$, suggesting that softly emphasizing the most relevant
successful chunks while still aggregating multiple candidates yields a more stable prior.

Finally, the success threshold $\eta$ controls the quality of online memory construction. A
smaller threshold allows more trajectories to be written into memory, but may introduce noisy or
partially failed segments. In contrast, an overly strict threshold improves memory precision but
slows down memory growth and reduces retrieval coverage. The default value $\eta=0.95$
provides a stable balance between memory quality and memory availability.

For adaptive prior guidance, we use the default configuration in
Table~\ref{tab:adaptive_t0_params}. We set $s_{\rm ref}$ to the same value as the retrieval
threshold $\gamma_{\rm sim}$, since both measure whether a retrieved candidate is sufficiently
close to the current observation. The scale $s_{\rm scale}=5\times10^{-4}$ is used to magnify
small differences in high cosine-similarity regimes. The clipping range $c_{\max}=50$ prevents
extreme confidence values from dominating the mapping. The coefficients $\alpha=1.0$ and
$\beta=0.05$ balance state-level similarity and action-level dispersion, while $\gamma=2.0$
controls the sharpness of the confidence-to-guidance mapping. We find that these parameters are
not sensitive within reasonable ranges, and the default setting works consistently across tasks.

Overall, these results suggest that the performance gain does not depend on a narrowly tuned
hyperparameter choice, and the default configuration provides a stable trade-off between retrieval
coverage, memory quality, and action-prior robustness.

\section{Visualization}

Additional qualitative results, including more videos and real-world robot demonstrations, are available at the following anonymous link:
\url{https://anonymous.4open.science/r/nips2026-1D23/}

\section{Limitations and Future Work}
Although our retrieve-then-steer framework improves the reliability of frozen generative VLAs, it still has several limitations. 
First, the method is designed for persistent deployment in relatively stable or slowly changing environments. 
Its effectiveness depends on the existence of reusable cross-episode experience; when object layouts, camera viewpoints, task goals, or robot calibration change rapidly, previously stored action priors may become less informative or even misleading. 
Second, the method requires the base policy to occasionally produce successful trials so that the online memory can be initialized and expanded. 
If the frozen VLA is far from competent on a target task, the memory may grow slowly and provide limited benefit.

Third, memory quality depends on the reliability of the progress or success estimator. 
Although our progress-calibrated prefix selection is designed to avoid storing failed, regressive, or post-success redundant segments, inaccurate progress estimation may still introduce noisy entries or discard useful successful experience. 
This issue is especially important because a small number of false successful entries can contaminate the retrieved prior and affect later generations. 
Fourth, retrieval based on visual similarity and trajectory consistency cannot fully resolve state aliasing. 
Visually similar states may require different actions due to subtle differences in object pose, contact state, occlusion, or task phase, which may lead to mismatched action priors even after similarity gating and consistency filtering.

Finally, our current memory management uses simple bounded storage and replacement strategies. 
While this is sufficient for the evaluated settings, larger-scale deployment may require more structured memory organization, long-term forgetting, task-aware indexing, and mechanisms for detecting environment changes. 
Future work will explore more reliable success verification, uncertainty-aware retrieval, scalable memory management, and adaptation to more dynamic environments with changing tasks and layouts.


\end{document}